\documentclass[sigconf]{acmart}

\usepackage{booktabs} 
\usepackage{caption}
\usepackage{subcaption}

\setcopyright{acmcopyright}

\copyrightyear{2024}
\acmYear{2024}
\setcopyright{rightsretained}
\acmConference[SAC '24]{The 39th ACM/SIGAPP Symposium on Applied
Computing}{April 8--12, 2024}{Avila, Spain}
\acmBooktitle{The 39th ACM/SIGAPP Symposium on Applied Computing (SAC
'24), April 8--12, 2024, Avila, Spain}
\acmDOI{10.1145/3605098.3635929}
\acmISBN{979-8-4007-0243-3/24/04}

\begin{document}
\title{Toward Scalable and Transparent Multimodal Analytics to Study Standard Medical Procedures: Linking Hand Movement, Proximity, and Gaze Data}
  
\renewcommand{\shorttitle}{Linking hand movement, proximity, and gaze data}

\author{Ville Heilala}
\authornote{Corresponding author}
\orcid{0000-0003-2068-2777}
\affiliation{
  \institution{University of Jyväskylä}
  \streetaddress{P.O. Box 35}
  \city{Jyväskylä} 
  \country{Finland}
  \postcode{40014}  
}

\author{Sami Lehesvuori}
\orcid{0000-0003-3889-5279}
\affiliation{
  \institution{University of Jyväskylä}
  \streetaddress{P.O. Box 35}
  \city{Jyväskylä} 
  \country{Finland}
  \postcode{40014}  
}

\author{Raija H{\"a}m{\"a}l{\"a}inen}
\orcid{0000-0002-3248-9619}
\affiliation{
  \institution{University of Jyväskylä}
  \streetaddress{P.O. Box 35}
  \city{Jyväskylä} 
  \country{Finland}
  \postcode{40014}  
}

\author{Tommi K{\"a}rkk{\"a}inen}
\orcid{0000-0003-0327-1167}
\affiliation{
  \institution{University of Jyväskylä}
  \streetaddress{P.O. Box 35}
  \city{Jyväskylä} 
  \country{Finland}
  \postcode{40014}  
}

\renewcommand{\shortauthors}{Heilala et al.}

\begin{abstract}
This study employed multimodal learning analytics (MMLA) to analyze behavioral dynamics during the ABCDE procedure in nursing education, focusing on gaze entropy, hand movement velocities, and proximity measures. Utilizing accelerometers and eye-tracking techniques, behaviorgrams were generated to depict various procedural phases. Results identified four primary phases characterized by distinct patterns of visual attention, hand movements, and proximity to the patient or instruments. The findings suggest that MMLA can offer valuable insights into procedural competence in medical education. This research underscores the potential of MMLA to provide detailed, objective evaluations of clinical procedures and their inherent complexities.
\end{abstract}

%
%



\maketitle

\section{Introduction}

Measurement and data collection instruments structure how we gather research data, whereas models and theories structure how we define what qualifies as valuable information \citep{Winne2019-wj}. Once integrated into scientific practice, the instruments inspire new theoretical concepts and pave the way for their acceptance within the scientific community \citep{Gigerenzer1991-iu}. Learning analytics (LA) involves collecting and analyzing educational data to understand better and improve learning \citep{Conole2011-zv} and multimodal multichannel trace data is suggested to hold promising potential in providing richer insights into domains of learning across various educational settings \citep{Azevedo2019-wo}. However, learning and its processes are complex. Thus, the more comprehensive and transparent data on learners, environments, and interactions can be traced \citep{worsley2021new}, the better possibilities of analyses and utilizations emerge. Utilizing diverse forms and sources of data, in other words, multiple data modalities \citep{Blikstein2013-mc} can enhance the precision and scope of understanding learner behaviors in their contexts \citep{di2018signals}. 

Despite the potential for educational research, multimodal and multichannel data collection presents methodological challenges such as instrumentation errors, lack of accuracy and replicability, handling data with varying dimensions (e.g., sampling rates, temporal alignment), securing internal and external validity, and ensuring the reliability of measures \citep{Azevedo2019-wo,Yan2022-vo}. Also, when collecting such data, a major issue with many commercial and proprietary measurement systems is the lack of financial scalability, methodological transparency, and control over the underlying algorithms used for data collection and analysis, which can lead to questions about the reliability, validity, and ethics \citep{David2022-eg,Yan2022-vo}. Open-source technologies and accessible APIs of hardware instruments can provide promising approaches for constructing scalable and transparent measurement systems \citep[e.g.,][]{Li2021-fk}; however, the systems are often in prototyping stages and might sacrifice accuracy or portability for affordability \citep{martinez2023lessons}.

This study represents a case of exploratory experimentation \citep{Devezer2021,Steinle1997-jw,Burian1997-pj} aiming to construct instrumentation for multimodal measurement and analysis of behavior in the context of nurse education. Efficient teamwork is essential in health care, and multimodal approaches to analyze complex dynamical behavior could provide insight, for example, into collaborative practices between health care professionals in educational settings and the field \citep{Zhao2023-lx}. Specifically, this study describes a minimum viable experiment (MVE) \citep{Devezer2021} to discover regularities concerning the complex dynamical behavior of a person conducting a medical ABCDE examination procedure (see, Section \ref{section:abcde}). The research aims to answer the following research question: What elements of the ABCDE procedure can be reconstructed from the multimodal hand movement, proximity, and gaze data by mainly utilizing affordable technology?

\section{Background}

\subsection{Multimodal learning analytics}

Multimodal learning analytics involves collecting, synchronizing, and analyzing various high-frequency data sources like video, logs, audio, and biosensors to study learning in various settings \citep{Blikstein2013-mc}. Different kinds of multimodal multichannel data streams are the key ingredients of MMLA, and \citet{Molenaar2023-ug} categorized them as physiological, behavioral, and contextual data. Physiological data, such as heart rate (HR) and electrodermal activity (EDA), have been associated with, for example, cognitive load management \citep[e.g.,][]{Lamsa2023-al} and emotional states \citep[e.g.,][]{Yadegaridehkordi2019-sz}. Behavioral data obtained, for example, using eye tracking and wearable motion detectors, can capture aspects of learners' activities like movement accuracy and situation awareness as they engage with learning content \citep[e.g.,]{Morita2022-kq,Cha2022-mq}. Contextual data like video recordings, positioning data, and self-reported measures offer insights into learners' interactions and experiences within various environments and learning situations \citep[e.g.,][]{Tokuno2023-fj,Fernandez-Nieto2021-fp,Heilala2022-xt}.

MMLA is expected to produce relevant, accountable, and actionable representations and interpretations while respecting the privacy of the stakeholders \citep{di2018signals,alwahaby2022evidence}. For this purpose, the use of interpretable and hyperparameter-free predictive models can produce minimal overhead, maximizing methodological transparency; an example of such a method is the \emph{Extreme Minimal Learning Machine} (EMLM) with full data as reference points and \emph{Mean-Absolute-Sensitivity} (MAS) to estimate the feature importances \citep{Linja2023feature}. \citet{ouhaichi2023research} review concluded four trending themes of the MMLA research: 1) addressing different contexts of learning, 2) focusing on self-regulated and collaborative learning processes, 3) encapsulation of multisensory affections from heterogeneous data, and 4) use of modern tools and methods for data analysis. Specifically, MMLA research conducted in real educational contexts, in other words, in the wild, is suggested to hold the potential for providing personalized learning experiences \citep{martinez2023lessons}. Regarding MMLA, this research aims to integrate behavioral data from hand movement, proximity, and eye-tracking utilizing scalable and transparent approaches to facilitate research of collaborative learning processes in the wild.

\subsection{Behavioral dynamics in health care}

Eye-tracking has become an instrumental tool in the medical field, for example, understanding cognitive load and assessing practitioner efficiency. \citet{Tokuno2023-fj} reviewed cognitive load assessment tools in surgical education, revealing a range of subjective and objective measures. Subjective tools included questionnaires like the NASA Task Load Index (NASA-TLX), while objective measures encompassed physiological parameters like heart rate variability, gaze entropy, gaze velocity, and pupil size. Also, gaze metrics have been utilized to assess non-technical skills like situation awareness in health care \citep[e.g.,][]{Cha2022-mq}. \citet{Ahmadi2022-wt} evaluated the mental workloads of Intensive Care Unit (ICU) nurses during their 12-hour shifts, focusing on how stress impacts their eye movement metrics. Their results suggested that periods of high stress seem to be associated with increased eye fixations and gaze entropy and decreased saccade duration and pupil diameter. \citet{Andrew_Wright2022-yw} utilized mobile eye-tracking to analyze visual attention patterns during an ultrasound-guided anesthesia procedure to differentiate between proficiency levels of practitioners. Their results showed that experienced medical professionals had fewer visual fixations, spent less time on the procedure, and exhibited less visual entropy, suggesting that eye-tracking can offer objective measures for assessing procedural competence and distinguishing expertise levels.

Proxemics--the study of how humans perceive and use space--and examinations of body movements can provide useful information on behaviors in medical education. Already \citet{Momen2010-jt} utilized a wireless Sony game controller's hardware, including a 3-axis accelerometer, to identify six nursing activities around a patient to improve hand hygiene prompts. By attaching five sensors to a nurse's body and analyzing the movements, the research found that the 1-Nearest Neighbour classifier was the most effective in identifying the activities. \citet{Morita2022-kq} used Bluetooth accelerometers and optical position tracking to examine microsurgical technical skills. \citet{Fernandez-Nieto2021-fp} pointed out the importance of spatial abilities in nursing, especially in effective team interactions and clinical procedures. Using indoor positioning sensors, their research transformed raw positioning data from nursing education classes into meaningful proxemics constructs like co-presence in interactional spaces, socio-spatial formations, and presence in spaces of interest with the aim of facilitating nurses' reflection, learning, and professional development in simulation-based training. However, indoor positioning systems often require stationary installations bound to a specific space.

Overall, the multimodal analytical advancements in medical education emphasize the importance of real-time assessment and its challenges. \citet{Cloude2022-zk} considered metacognition and self-regulation in clinical reasoning and argued that medical education faces challenges in effectively analyzing learning during activities, as most educational settings utilize intermittent assessments that miss real-time information on knowledge, skills, and abilities, highlighting a need for approaches like MMLA. Furthermore, based on an MMLA implementation in nursing education, \citet{martinez2023lessons} pointed out that MMLA systems need to be trustworthy and address data incompleteness while balancing high-quality data capture with portability and affordability of sensors and consider users' concerns about potential distractions and inconvenience due to being monitored.

\subsection{The ABCDE approach}\label{section:abcde}

Healthcare professionals utilize various standard procedures when diagnosing patients. In this study, we focus on one such procedure that goes by the acronym ABCDE, which stands for Airway, Breathing, Circulation, Disability, and Exposure. It refers to a systematic protocol primarily used in emergency medicine but applies to other healthcare areas \citep{Thim2012-cg}. It serves as a universal approach for patient assessment and directs medical professionals, particularly nurses, in conducting an efficient and comprehensive assessment of a critically ill patient's condition \citep{Schoeber2022-og}. The completion of the assessment involves five stages consisting of different simultaneous and continuous assessment and treatment steps \citep{Thim2012-cg}. The procedure starts with an airway assessment to ensure the patient has a clear breathing passage. The patient's respiratory rate and quality are then examined during the breathing analysis. The patient's blood pressure and heart condition are evaluated during the circulation inspection. In the disability stage, neurological function is examined, typically through a quick assessment of the patient's responses. The final step involves a prompt but thorough examination of the patient's body to look for any additional symptoms of disease or trauma. The main aim of the ABCDE approach is that healthcare professionals can accurately prioritize treatments and interventions by consistently following a protocol that simplifies complex clinical situations, allowing them to establish common situational awareness among the medical team and save valuable time \citep{Thim2012-cg}. 

Despite the wide use of the ABCDE approach in various clinical settings, \citet{Schoeber2022-og} found that healthcare professionals' theoretical knowledge of the approach varies based on the professionals' type of department, profession category, and age. The result suggests a need to more closely examine the underlying individual differences beyond theoretical knowledge. For example, eye-tracking has been successfully used in evaluating the medical professionals' performance in the ABCDE approach. \citet{Fernandez-Mendez2019-on} utilized eye-tracking to study how lifeguards performed the ABCDE approach. They found that the lifeguards' performance was misaligned with multimodal data: none of the lifeguards completed the approach correctly, but most of their visual fixations during the assessment procedure were shared between the essential areas for the approach, indicating that eye-tracking could be a valuable method for evaluating the performance of medical procedures. \citet{Lee2019-cq} utilized eye-tracking, log data, and self-reported cognitive load measurements to assess the performance of the ABCDE approach between experts and novices in a medical simulation game. Their results indicated that experts outperformed novices regarding speed, accuracy, and cognitive load, associated with higher prior knowledge. 

\section{Materials and methods}

\subsection{Experimental setting}

Two nurse educators specialized in critical care conducted the ABCDE procedure on an actor patient. The experiment was conducted in a classroom, simulating a real medical examination room with real medical equipment and a hospital bed (Figure 1). An actor patient played the role of a patient who had arrived from an appendectomy, a common surgery operation. The task of the participating nurses was to conduct the ABCDE procedure for evaluating the patient's condition. The participant was required to work close to the patient's bed and utilize the instrument table positioned 6 meters away from the center of the hospital bed. Multimodal measurement was used to record the participants' hand movements, gaze dynamics, and proximity data. Both participants performed the procedure twice, and the data were collected for the initial and repeated experiments.

\begin{figure}[htb!]
    \centering
    \includegraphics[width=.8\linewidth]{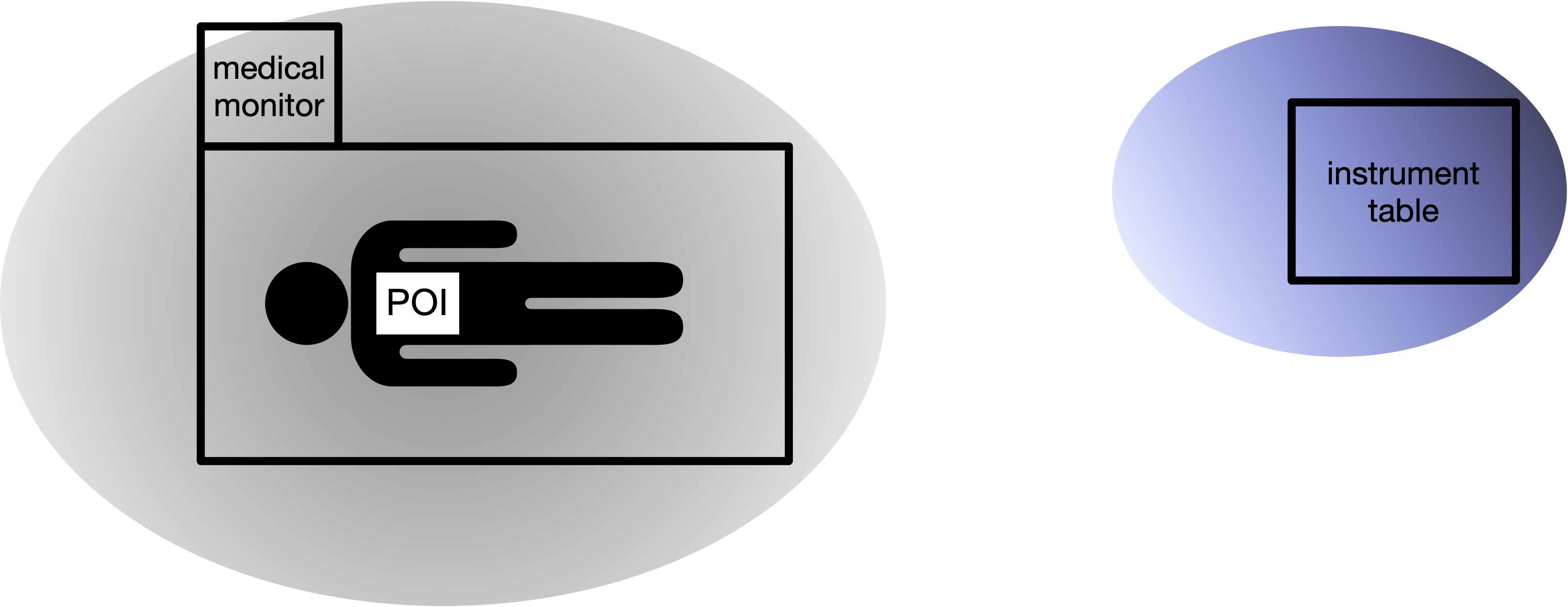}
    \caption{Experimental setting where the point of interest (POI) indicates the reference point for proximity estimation. The gray area indicates an area close to the patient, whereas the blue area is a specific area far from the patient.}
    \label{fig:experiment}
\end{figure}

\subsection{Apparatus}

The measurement system (Figure 2) consisted of wireless and wired sensors and recording devices connected to a Raspberry Pi 4 (8 GB) microcomputer that served as a hub for collecting and synchronizing the data streams and forwarding them to a recording laptop through the Lab Streaming Layer (LSL). The system's architectural design is aimed at being extendable for adding additional measurement instruments and scalable for measuring multiple subjects. Apart from the eye-tracking device, the other devices were relatively affordable and accessible and utilized open-source technologies. In this study, the system was capable of real-time measurement and synchronization of five data modalities: hand movements using wireless accelerometers, proximity estimation based on Bluetooth Low Energy (BLE) signal, eye tracking using Tobii Pro Glasses 3, video recording and discrete markers used for real-time annotation. Markers were used as a reference point for evaluating the latency of each individual measurement device. In general, the highest latency of the system was assessed to be approximately 50 ms. Wearable accelerometers and the eye tracker were wireless, allowing free and safe movement of the participants.

\begin{figure}[htb!]
    \centering
    \includegraphics[width=.8\linewidth]{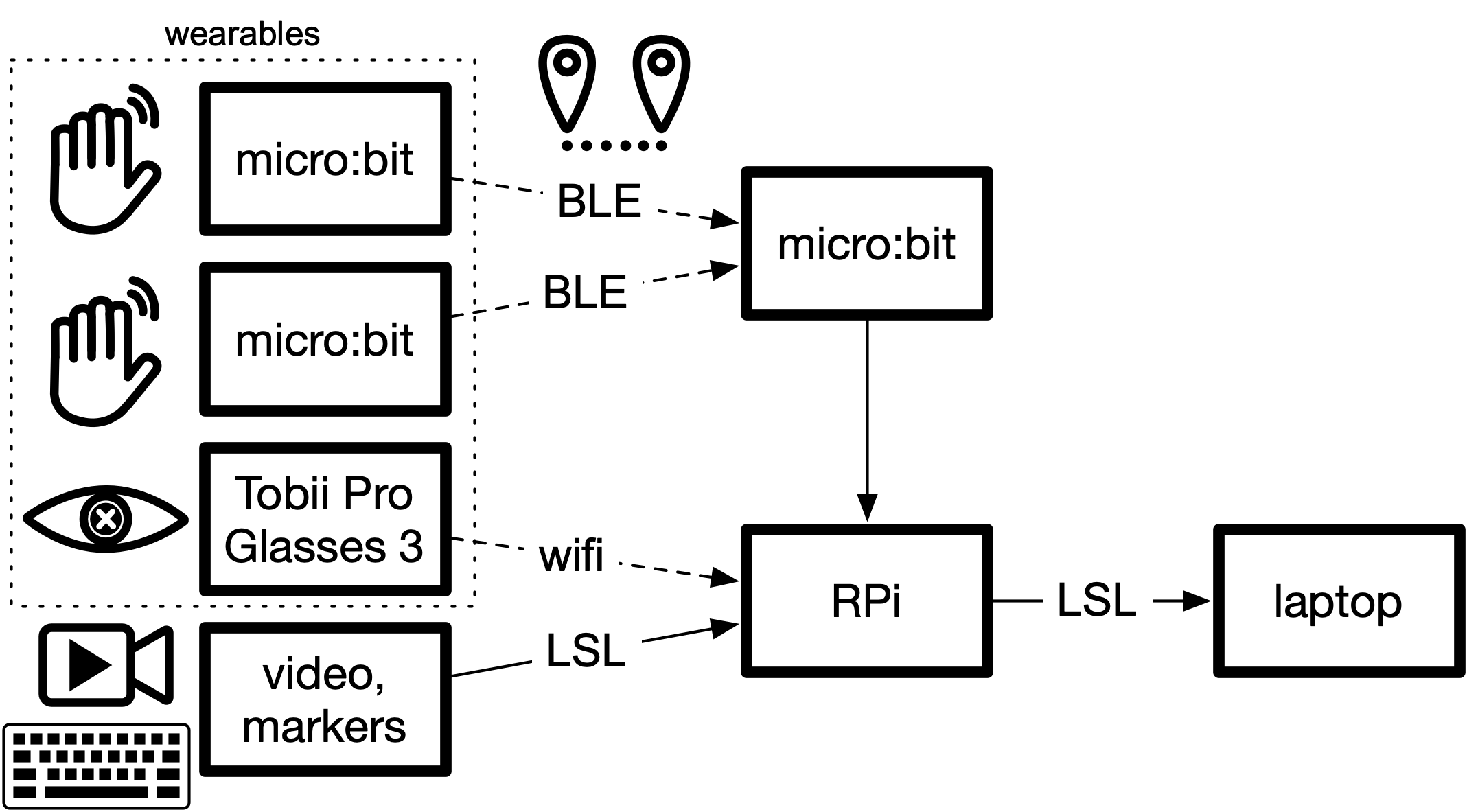}
    \caption{An overview of the apparatus used to measure and record all five data streams.}
    \label{fig:apparatus}
\end{figure}

\paragraph{Hand movement}

The micro:bit is a small, versatile, affordable, and programmable open-source ARM-based microcontroller intended for educational and learning purposes, focusing on teaching children the fundamentals of programming and electronics. It includes a variety of sensors, input and output options, and an environment for block-based programming. For example, the micro:bit contains a built-in 3-axis accelerometer that can detect motion, orientation, and tilt. Promoting the constructionist approach by encouraging building interactive projects and engaging with technology, the micro:bit facilitates hands-on learning and fosters creativity and problem-solving skills (Austin et al., 2020). Micro:bit has been used in several studies relating to computing education (e.g., Andersen, 2022). However, to our knowledge, it has not been utilized as a measurement device for scientific work. This study aimed to pilot and evaluate micro:bit as a scientific instrument. Thus, two micro:bit devices were connected to an add-on shield to enable battery power and wireless wristband use. The devices were attached to the wrists of the participants. The built-in 3-axis accelerometer measured hand movements with a sampling rate of 40 Hz. Devices sent the raw accelerometer signal values using BLE connection to a third micro:bit connected to the RPi receiver.

\paragraph{Proximity}

Spatial behavior in terms of proximity was measured using the Received Signal Strength Indicator (RSSI) of the accelerometers, which were connected using BLE to the third micro:bit serving as a receiver. RSSI in Bluetooth technology is a metric that quantifies the power level of a received radio signal. It is commonly used to estimate the distance between devices, as signal strength typically decreases with increasing distance. By employing wireless Bluetooth-based instruments and RSSI, researchers can utilize proximity-based methods \citep[e.g.,][]{MartinezMaldonado2020-pa}. RSSI values can be influenced by various factors, such as environmental conditions and obstacles that interfere with radio waves \citep[e.g.,][]{Kim2015-qa}. In this study, no significant structures were interfering with the Bluetooth signal. Thus, the raw RSSI signal values between hand movement sensors and the receiver were used, and the signal was calibrated based on the closest and farthest distance to the point of interest (POI). The third receiver micro:bit was placed on the chest of the actor patient, serving as the POI (Figure 1). The farthest point was chosen to be the table containing some of the medical instruments the participants had to use in the procedure. The experiment was designed spatially so that the participant moved mainly around the patient's bed and between the bed and the medical instrument table. 

\paragraph{Gaze}

\begin{figure}[htb!]
    \centering
    \includegraphics[width=.8\linewidth]{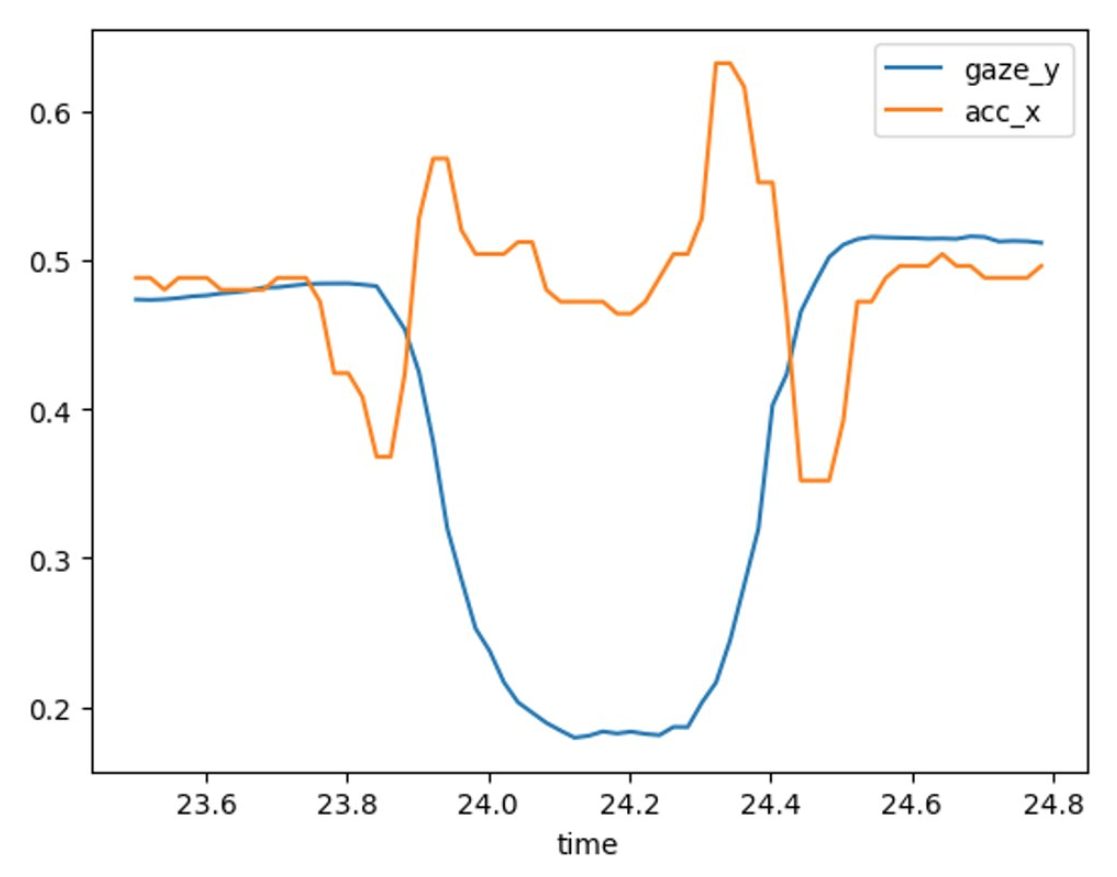}
    \caption{Synchronized head movement (acc\_x) and eye movement (gaze\_y)}
    \label{fig:sync}
\end{figure}

Tobii Pro Glasses 3 eye tracking device (sampling rate 50 Hz) was used to record participants' eye movements. The raw signal was the (x, y) coordinate of the participants' gaze in the visual measurement plane of the device. The coordinate values were continuous and in [0, 1]. Blinks were coded as missing values because they caused interruptions in the measurement signal. Tobii Pro Glasses 3 API was used to communicate with the eye-tracking device. The synchronization of the accelerometer and eye-tracking signals was verified by asking the subject to fixate gaze on a stationary point and perform a slow vertical head movement while the micro:bit was attached to the forehead of the subject wearing the eye-tracking glasses. For example, a similar approach has been used to synchronize eye-tracking and motion-capture systems \citep{Burger2018-pm}. Figure \ref{fig:sync} shows the synchronized vertical head movement (up and down) measured using the micro:bit and the slowly changing vertical eye movement when fixating on a stationary point. The use of raw gaze signals in this study allowed context-free analysis without the need to define areas of interest (AOI).

\paragraph{Video and markers}

The video recorder setup consisted of a laptop and a webcam. The webcam stored the raw video file on the laptop's local hard drive and sent the video stream's frame numbers over LSL to the recording laptop. Synchronized frame numbers enabled the synchronization of the video with other multimodal data. Also, the laptop was used to send and synchronize keyboard markers over LSL for live annotation of the experiment. Markers were used to sequence the multimodal data according to the steps and phases of the ABCDE procedure. Synchronized video and markers served as ground truth for validating the analysis results. 

\subsection{Analysis procedure}

To reduce the incoherence in the RSSI signal, proximity was operationalized based on the strongest signal of the accelerometers in the right hand (rh) and left hand (lh) for each time point \textit{t}, ${RSSI}^{t}_{max}$ = ${max}({RSSI}^{t}_{rh}, {RSSI}^{t}_{rh})$. Based on the calibration measurements in the experiment, the signal was discretized as a binary variable to indicate the time points when the participant was working beside the patient and beside the medical instrument table. A missing value suggested that the participant was located somewhere in the intermediate space. Hand movements were operationalized using the velocity of the movement. Before calculating velocity, the signal was preprocessed by applying a Savitzky–Golay filter for denoising \citep{Schafer2011-nx,Karaim2019-yn}.

Entropy provides a useful metric for understanding the degree of variability, disorder, or unpredictability in the studied data or system. For example, entropy has been used to examine the development of attention to faces \citep{Frank2009-sn} and webpage aesthetics \citep{Gu2021-nh}. Stationary gaze entropy reflecting the overall spatial dispersion of gaze was used to operationalize gaze dynamics between explorative (i.e., wider gaze dispersion) and exploitative (i.e., limited gaze dispersion) phases where lower entropy is interpreted to indicate more exploitative, spatially focused, and coherent visual focus \citep{Shiferaw2019-ld,Holmqvist2011-tr,Frank2009-sn,Lanini-Maggi2021-vj}.

In the context of information theory, entropy quantifies the uncertainty or randomness of a set of outcomes or events. Entropy can be quantified using the Shannon entropy \citep{Shannon1948-gy}, defined as the average Shannon information content of an outcome \citep{MacKay2003-qt}. In other words, it quantifies the average amount of information needed to describe an outcome from a random variable following a given probability distribution. Measured gaze data concerns two coordinate variables. Using the logarithm base 2, Shannon entropy is measured in bits \citep{Shannon1948-gy} and the joint entropy of two variables is \citep{MacKay2003-qt}: $H(x,y) = \sum_{x\in \mathcal{X},y\in \mathcal{Y}}P(x,y)\log_2\frac{1}{P(x,y)}$.

Raw gaze signal was preprocessed using cubic spline imputation \citep[e.g.,][]{Frank2009-sn} to deal with the missing coordinate values caused by blinks. A probability distribution of the continuous gaze measurement signal was needed for calculating the joint entropy. To create a probability distribution of the continuous gaze measurement signal, the data was discretized into equally sized bins representing the state space of gaze behavior. In other words, the discretization divided the measurement plane of the eye tracker as a 100 x 100 matrix, each cell depicting the probability of gazing at that section of the visual plane during a specific time period. Entropy was calculated for a sliding window of 5 seconds. To evaluate the robustness of the approach, different discretization group sizes (i.e., 10, 25, 50, 75) and sliding windows (i.e., 2, 3, 4, 6) were tested. However, the results were qualitatively the same.

\begin{figure*}[htb!]
    \centering
    \includegraphics[width=.65\textwidth]{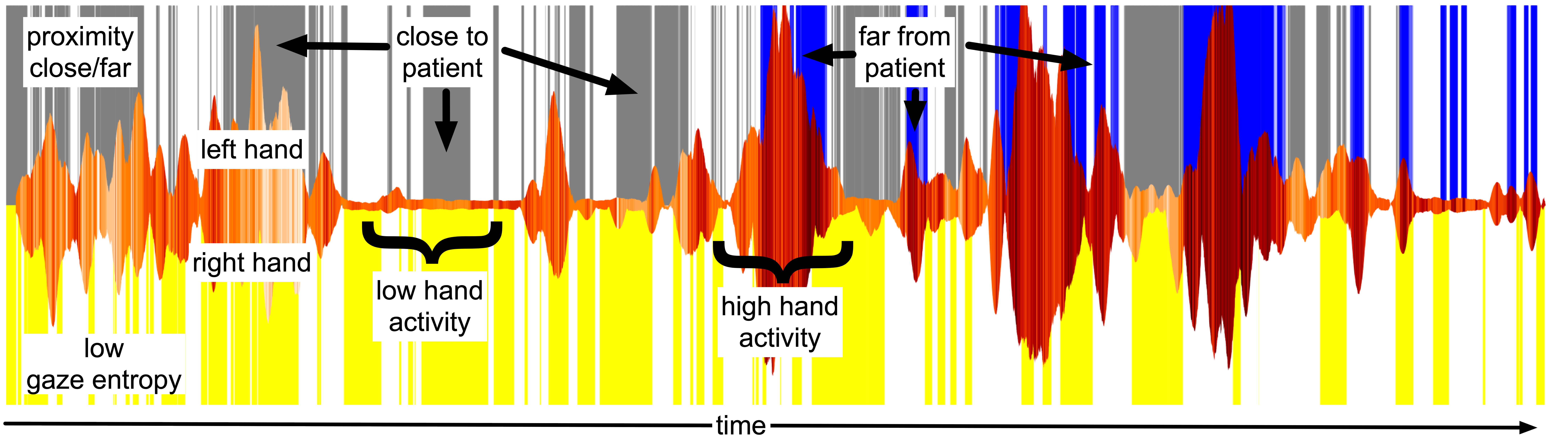}
    \caption{An example of a behaviorgram fusing the dynamics of the multimodal data}
    \label{fig:behaviorgram}
\end{figure*}

The measurements were visualized using a behaviorgram, a graphical representation that visually depicts patterns of behavior, interactions, or activities over time. Like exploratory data analysis, visual analytics aims to uncover knowledge and acquire insight from complex data sets \citep{Cui2019-os}. Behaviorgrams can be used in visual analytics to understand the behavior of individuals or groups in contexts such as psychology and human-computer interaction \citep[e.g.,][]{Chen2019-nb}. The custom extended behaviorgram (Figure \ref{fig:behaviorgram}) presented in this study exploits dimensional stacking and the dense pixel technique \citep{Keim2002-jt,Keim2001-og} to visualize temporal relationships of all the measured dimensions.

The central axis of the behaviorgram represents temporal hand movement velocities as an accelerograph. The accelerograph is asymmetrical concerning the central line, the lower part representing the right hand and the upper part representing the left hand. Color coding of the accelerograph exploits the dense pixel technique, a sort of heatmap that depicts higher RSSI signal strength in a brighter color, indicating higher proximity to the POI. The accelerograph's upper temporal segment illustrates the participant's binary position (i.e., beside the patient, beside the instrument table). The lower segment depicts the gaze entropy, where the mean entropy was set as the threshold for marking a segment as denoting low entropy (i.e., more coherent and spatially focused visual perception). Furthermore, the extended behaviorgram was reduced to a more simplified behaviorgram (Figures \ref{fig:nurse1} and \ref{fig:nurse2}). The simplified behaviorgram captures proximity and combines the dimension of hand movement and gaze entropy, specifically illustrating the participant's behavioral dynamics concerning the patient. Behaviorgrams were discretized into broad behavioral phases based on video observation and marker annotations.

\section{Results}

\begin{figure*}[htb!]
    \centering
     \begin{subfigure}[b]{0.48\textwidth}
         \centering
         \includegraphics[width=\textwidth]{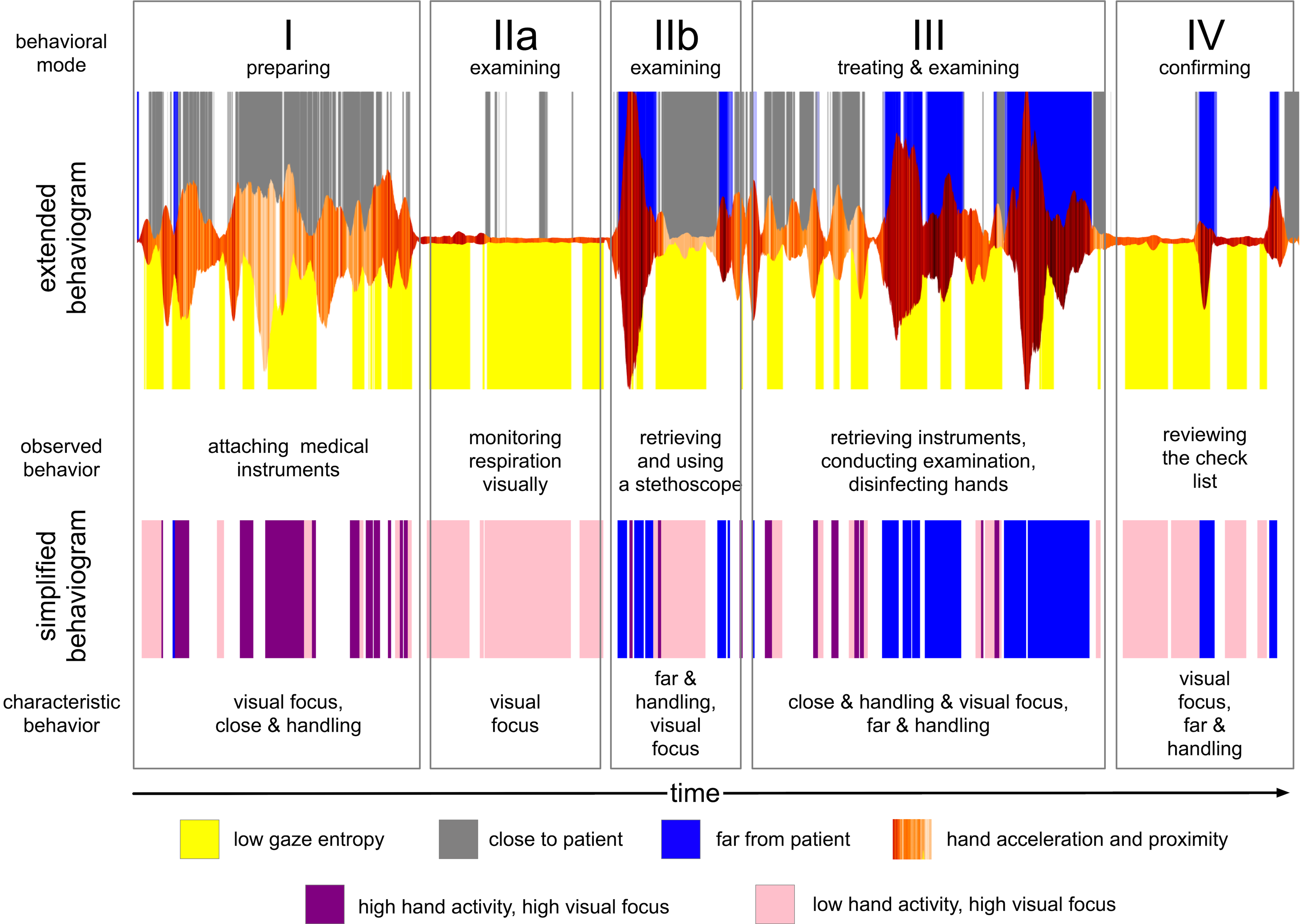}
         \caption{The initial examination}
         \label{fig:nurse11}
     \end{subfigure}
     \hfill
     \begin{subfigure}[b]{0.48\textwidth}
         \centering
         \includegraphics[width=\textwidth]{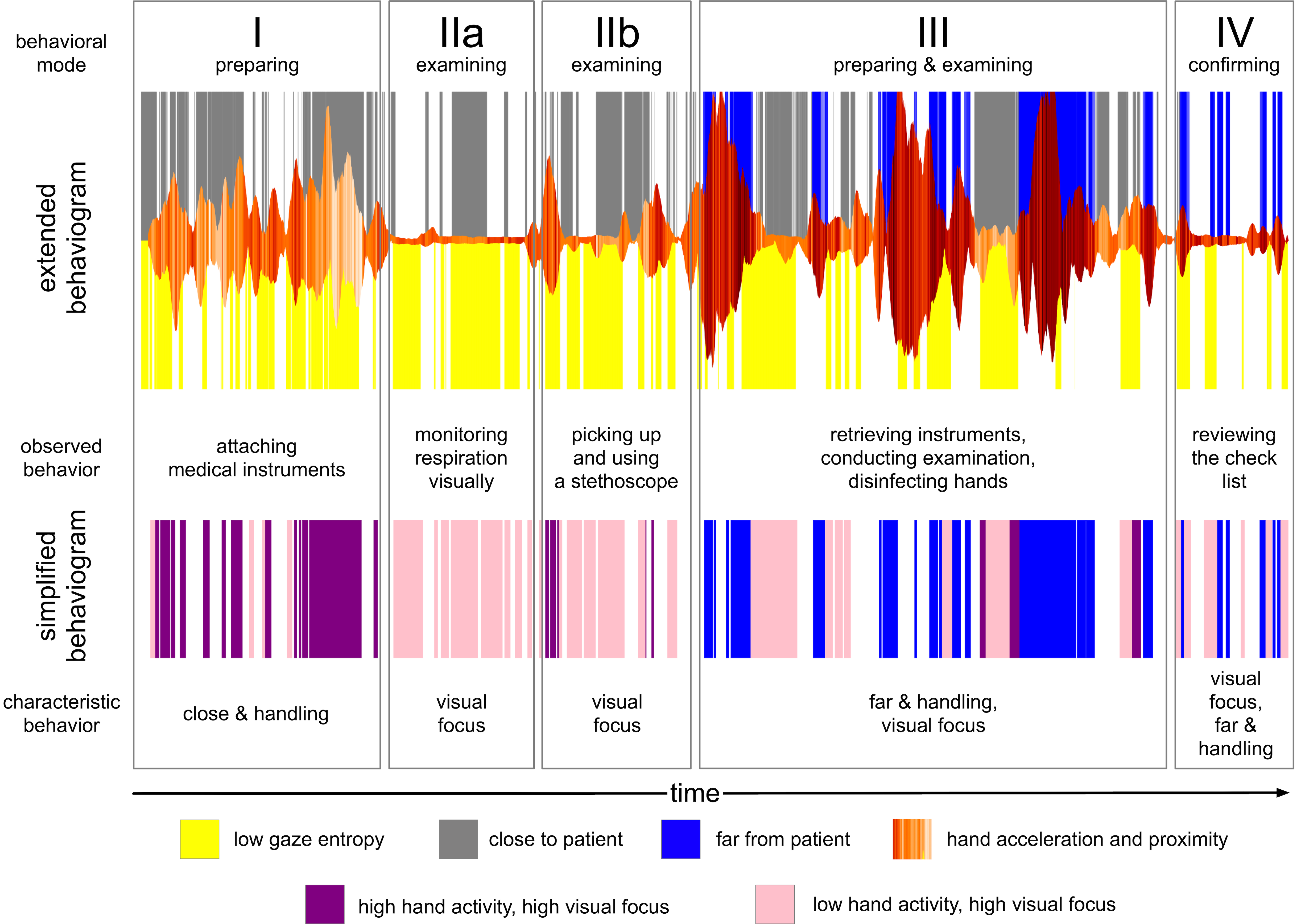}
         \caption{The repeated examination (as in Figure \ref{fig:behaviorgram})}
         \label{fig:nurse12}
     \end{subfigure}
\caption{Behaviorgrams of Nurse 1}
\label{fig:nurse1}
\end{figure*}

\begin{figure*}[htb!]
    \centering
     \begin{subfigure}[b]{0.49\textwidth}
         \centering
         \includegraphics[width=\textwidth]{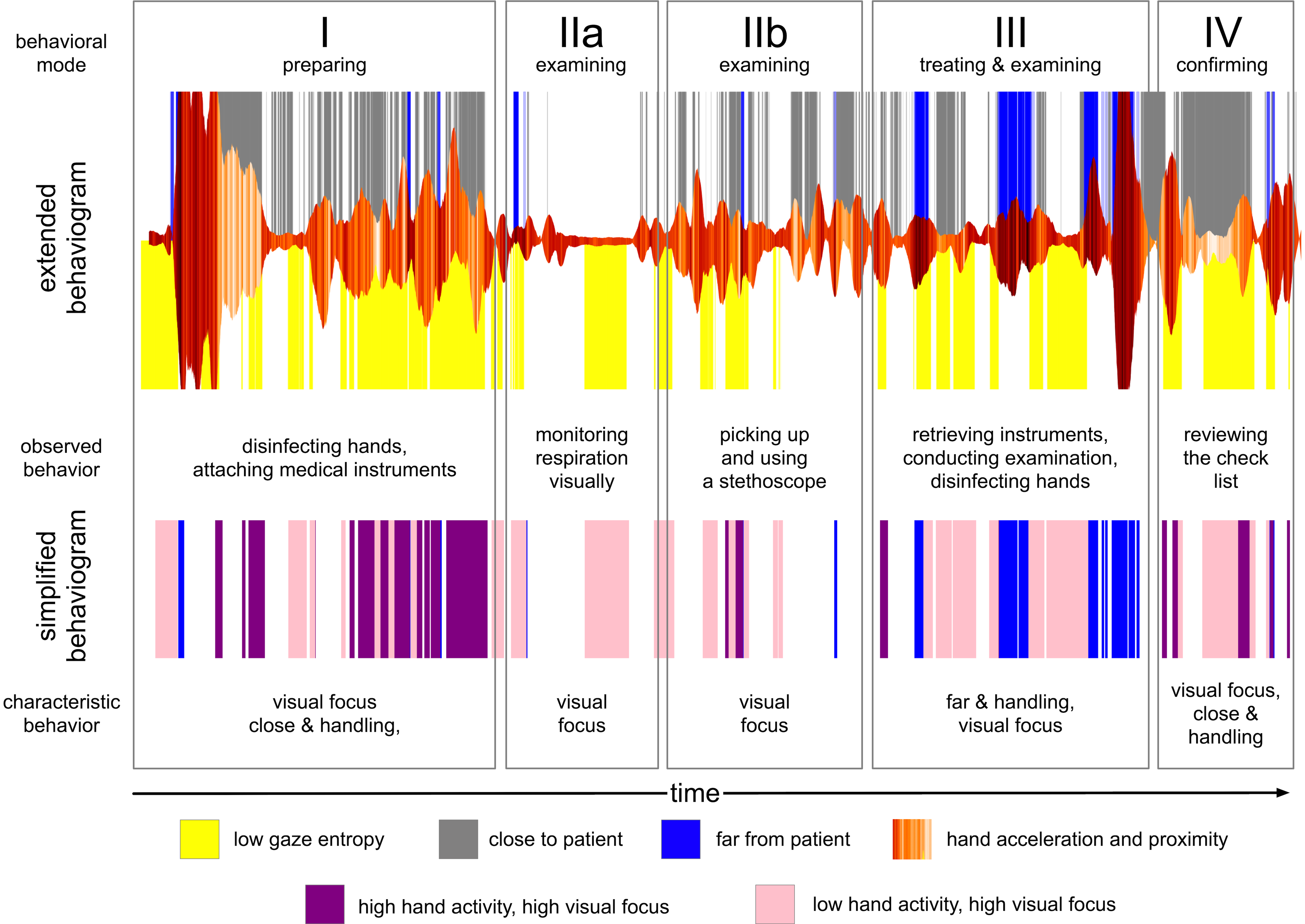}
         \caption{The initial examination}
         \label{fig:nurse21}
     \end{subfigure}
     \hfill
     \begin{subfigure}[b]{0.49\textwidth}
         \centering
         \includegraphics[width=\textwidth]{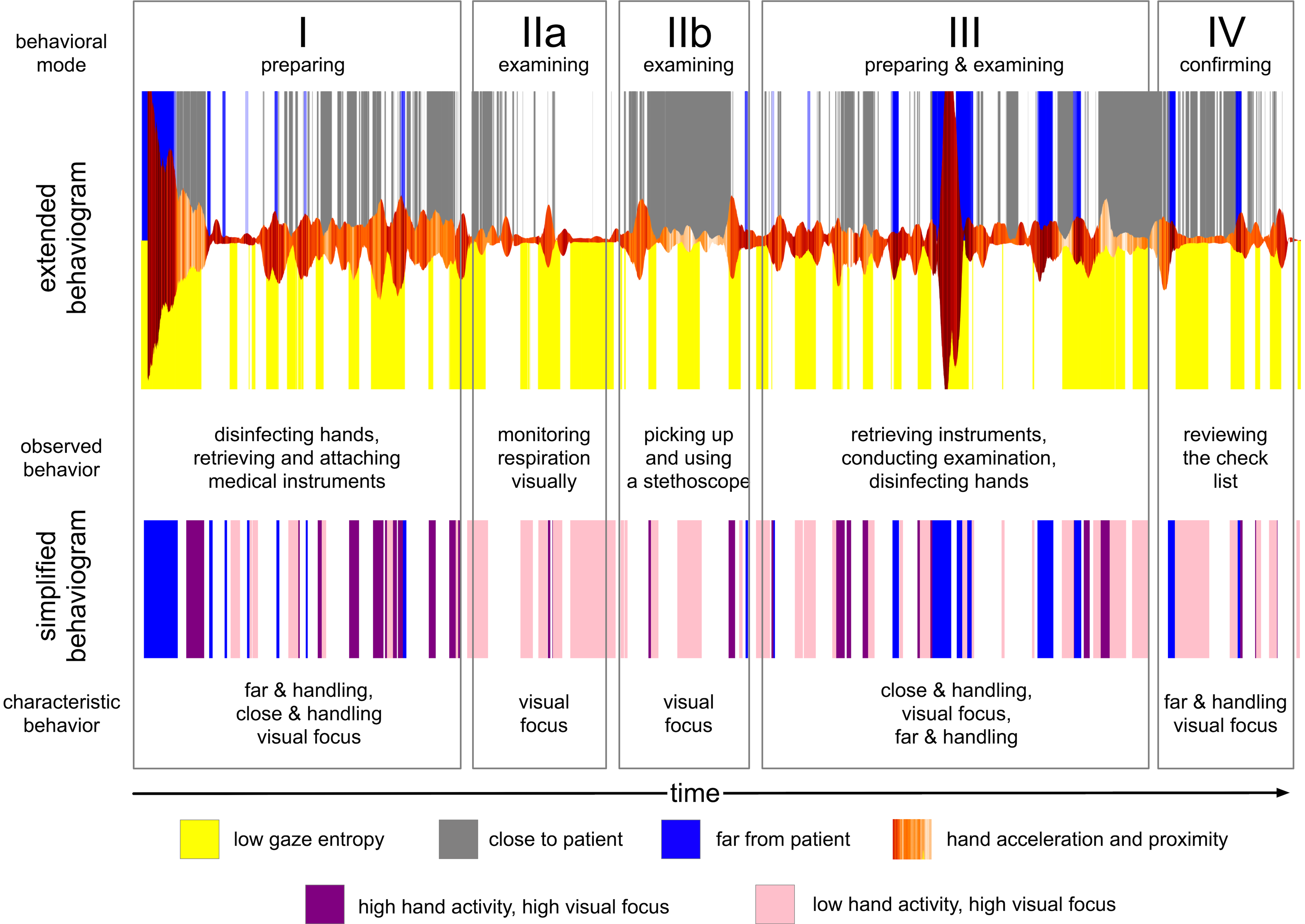}
         \caption{The repeated examination}
         \label{fig:nurse22}
     \end{subfigure}
\caption{Behaviorgrams of Nurse 2}
\label{fig:nurse2}
\end{figure*}

Based on visual analytics, the expected steps in the ABCDE procedure, and validation according to video recordings and annotations, behaviorgrams were found to consist of four phases (Figures \ref{fig:nurse1} and \ref{fig:nurse2}). In Phase I, the nurse retrieved medical instruments and attached them to the patient, which involved hand movements close to the patient while maintaining high visual focus when handling the instruments. Phase II consisted of monitoring respiration frequency by observing the patient's chest, confirming visually other vital signs from the medical monitor (IIa), and using a stethoscope to auscultate the patient's chest (IIb). In phase II, the nurses were positioned either close to the patient or in the intermediate space between the patient and the instrument table. Phase II corresponds to the Breathing assessment in the ABCDE approach, and it is characterized by high visual attention as measured using gaze entropy because the tasks require focusing on the patient and the monitor. Phase IIa involved only visual observations without hand movements, and in Phase IIb, some hand movements can be seen in the behaviorgram because of the use of a stethoscope.

The latter part of the ABCDE procedure (Phase III) involved fetching instruments from the instrument table and performing small examination operations close to the patient (e.g., measuring body temperature and giving medication). Thus, Phase III corresponds to the Circulation, Disability, and Exposure assessments in the ABCDE approach, and it is characterized by changes in proximity and alternating hand activity combined with low gaze entropy (i.e., coherent visual perception). Phase IV consisted of retrieving a checklist and reviewing the patient's condition according to the list. The phase was characterized by changes in proximity, a few short periods of hand movements, and visual focus.

Accelerographs representing hand movement velocity showed specific dynamical patterns based on the phases of the ABCDE procedure. The preparation (Phase I) involved attaching medical instruments to the patient, which is seen as a continuous period of high hand movements in all behaviorgrams. Specifically, hand disinfection was clearly shown as having high peaks in velocity and a higher distance from the patient because it was performed at the instrument table. For example, Nurse 1 performed three hand disinfections in Phase III in the repeated experiment (Figure \ref{fig:nurse12}). On the other hand, the phases where the nurse mainly observed the patient's condition visually were characterized by low hand activity (Phase IIa, IIb, and IV).

\section{Discussion and conclusion}

The results of this study underscore the potential of using multimodal learning analytics in understanding behavioral dynamics in the medical field. Utilizing relatively affordable technology and visual analytics, the research was able to trace the different phases of the ABCDE procedure and discern the behavioral patterns associated with each phase. The clear co-occurrence of hand movement activity, gaze entropy, and spatial location across various stages suggest that these metrics provide insights into the dynamics of the procedure. Notably, the low gaze entropy indicated periods of consistent visual perception throughout the procedure, suggesting that medical professionals frequently alternate between explorative and exploitative gazes, especially during intricate procedures. This is particularly significant when considering the importance of visual attention in medical tasks and how it can influence the outcome of procedures.

Integrating multimodal data into a behaviorgram revealed distinct visual patterns based on the different phases of the ABCDE procedure. The results showed that preparation for the procedure, breathing assessment, Circulation/Disability/Exposure phase, and review phase could be identified. Different periods of eye-hand coordination can be distinguished when combining gaze entropy with information from the accelerograph (i.e., high hand activity and high visual focus, low hand activity and high visual focus). Specifically, phases characterized mainly by visual observation displayed visual focus and reduced hand activity, thus allowing for the differentiation between manual and observational phases of the procedure. The proximity measure captured the movement of the nurse between the patient and the instrument table. In general, multimodal behaviorgrams and results based on visual analytics could be linked with actual behavioral dynamics during the procedure.

The results highlight that multimodal multichannel data collection approaches could and should be examined for validity before feeding the data to complex machine learning and artificial intelligence algorithms. Before engaging with more complex analysis techniques, it can be helpful to utilize visual analytics to examine the potential patterns in the data. Such an approach could assist in validating the measurement procedures and facilitating transparency of the more complex analyses. The results provided initial evidence of validity and reliability: results aligned with visual analytics and observed behavior in marker-annotated video recordings for both the initial and repeated examinations for both subjects. In other words, the results showed evidence of within-subject and between-subjects similarity. 

The multimodal multichannel measurement in this study utilized relatively affordable technology (i.e., Raspberry Pi, micro:bit), enabling the techniques to be scaled for multiple subjects. The results showed that micro:bit has the potential to produce accurate multimodal measurement data while Raspberry Pi functions as a recording device. The expensive part of the instrumentation was the eye-tracking device; however, more affordable devices are potentially being introduced to the market as technology advances. Scalable and transparent measurement and analysis of behavioral dynamics can enable research in the wild, which refers to approaches to studying and understanding human behavior and technology interactions in real-world, everyday settings, as opposed to controlled lab environments \citep{Hutchins1995-bp,martinez2023lessons}. For example, such approaches can enable research in medical situations where an observer can not enter the room of a patient \citep[e.g.,][]{Faiz2021-le}. Furthermore, \citet{Kolbe2019-ym} pointed out the limitations of traditional team research methods in healthcare, which often focus on static descriptions rather than dynamic team processes over time. They suggested that more profound insights into the intricacies of teamwork can be achieved by adopting methodological approaches that consider dynamics, such as event- and time-based observations, social sensor-based measurement, and micro-level coding. Thus, potential applications of the approach presented in this research could include the analysis of situation awareness, professional noticing, joint visual attention in collaborative tasks, understanding the dynamics of patient care, and exploring how medical instruments are handled in real-world scenarios. These insights could inform training programs, process improvements, and even technology design for healthcare contexts. However, it is worth being aware of and clearly defining the limits and scope of the multimodal approaches; in other words, ``noting one's paradigm's relatively well-marked perimeter is a hallmark of sound and responsible science'' \citep[][p.~288]{Winne2019-wj}. In conclusion, this research adds value to medical education by emphasizing the importance of integrating multimodal measures to understand medical professionals' behavior during standard procedures comprehensively.

\subsection{Limitations and future research}

Like other similar studies implementing an exploratory approach, this study can be criticized due to its lack of a control group and the generalisability of the results. Furthermore, while behaviorgram provides a comprehensive overview of behavior over time, it does not capture nuanced processes underlying specific actions. Finally, the study utilized a very small sample size, and the generalizability of the results to other medical procedures beyond the ABCDE approach remains to be explored. Future studies need to utilize more comprehensive analysis techniques and delve deeper into the individual differences among professionals and how these might influence the observed behaviors. Furthermore, the critical issue for future work is to examine how to bridge observed behavioral dynamics with cognitive functions and outcomes.

\begin{acks}
The work was supported by the Research Council of Finland under Grant number 353325.
\end{acks}

\bibliographystyle{ACM-Reference-Format}
\bibliography{refs} 


\begin{thebibliography}{53}


\ifx \showCODEN    \undefined \def \showCODEN     #1{\unskip}     \fi
\ifx \showDOI      \undefined \def \showDOI       #1{#1}\fi
\ifx \showISBNx    \undefined \def \showISBNx     #1{\unskip}     \fi
\ifx \showISBNxiii \undefined \def \showISBNxiii  #1{\unskip}     \fi
\ifx \showISSN     \undefined \def \showISSN      #1{\unskip}     \fi
\ifx \showLCCN     \undefined \def \showLCCN      #1{\unskip}     \fi
\ifx \shownote     \undefined \def \shownote      #1{#1}          \fi
\ifx \showarticletitle \undefined \def \showarticletitle #1{#1}   \fi
\ifx \showURL      \undefined \def \showURL       {\relax}        \fi
\providecommand\bibfield[2]{#2}
\providecommand\bibinfo[2]{#2}
\providecommand\natexlab[1]{#1}
\providecommand\showeprint[2][]{arXiv:#2}

\bibitem[Ahmadi et~al\mbox{.}(2022)]%
        {Ahmadi2022-wt}
\bibfield{author}{\bibinfo{person}{Nima Ahmadi}, \bibinfo{person}{Farzan Sasangohar}, \bibinfo{person}{Jing Yang}, \bibinfo{person}{Denny Yu}, \bibinfo{person}{Valerie Danesh}, \bibinfo{person}{Steven Klahn}, {and} \bibinfo{person}{Faisal Masud}.} \bibinfo{year}{2022}\natexlab{}.
\newblock \showarticletitle{Quantifying Workload and Stress in Intensive Care Unit Nurses: Preliminary Evaluation Using Continuous {Eye-Tracking}}.
\newblock \bibinfo{journal}{\emph{Human factors}} \bibinfo{volume}{0}, \bibinfo{number}{0} (\bibinfo{date}{May} \bibinfo{year}{2022}), \bibinfo{pages}{187208221085335}.
\newblock
\showISSN{0018-7208, 1547-8181}
\urldef\tempurl%
\url{https://doi.org/10.1177/00187208221085335}
\showDOI{\tempurl}


\bibitem[Alwahaby et~al\mbox{.}(2022)]%
        {alwahaby2022evidence}
\bibfield{author}{\bibinfo{person}{Haifa Alwahaby}, \bibinfo{person}{Mutlu Cukurova}, \bibinfo{person}{Zacharoula Papamitsiou}, {and} \bibinfo{person}{Michail Giannakos}.} \bibinfo{year}{2022}\natexlab{}.
\newblock \showarticletitle{The Evidence of Impact and Ethical Considerations of Multimodal Learning Analytics: A Systematic Literature Review}.
\newblock In \bibinfo{booktitle}{\emph{The Multimodal Learning Analytics Handbook}}, \bibfield{editor}{\bibinfo{person}{Michail Giannakos}, \bibinfo{person}{Daniel Spikol}, \bibinfo{person}{Daniele Di~Mitri}, \bibinfo{person}{Kshitij Sharma}, \bibinfo{person}{Xavier Ochoa}, {and} \bibinfo{person}{Rawad Hammad}} (Eds.). \bibinfo{publisher}{Springer International Publishing}, \bibinfo{address}{Cham}, \bibinfo{pages}{289--325}.
\newblock
\showISBNx{9783031080760}
\urldef\tempurl%
\url{https://doi.org/10.1007/978-3-031-08076-0\_12}
\showDOI{\tempurl}


\bibitem[Azevedo and Ga{\v s}evi{\'c}(2019)]%
        {Azevedo2019-wo}
\bibfield{author}{\bibinfo{person}{Roger Azevedo} {and} \bibinfo{person}{Dragan Ga{\v s}evi{\'c}}.} \bibinfo{year}{2019}\natexlab{}.
\newblock \showarticletitle{Analyzing Multimodal Multichannel Data about {Self-Regulated} Learning with Advanced Learning Technologies: Issues and Challenges}.
\newblock \bibinfo{journal}{\emph{Computers in human behavior}}  \bibinfo{volume}{96} (\bibinfo{date}{July} \bibinfo{year}{2019}), \bibinfo{pages}{207--210}.
\newblock
\showISSN{0747-5632}
\urldef\tempurl%
\url{https://doi.org/10.1016/j.chb.2019.03.025}
\showDOI{\tempurl}


\bibitem[Blikstein(2013)]%
        {Blikstein2013-mc}
\bibfield{author}{\bibinfo{person}{Paulo Blikstein}.} \bibinfo{year}{2013}\natexlab{}.
\newblock \showarticletitle{Multimodal learning analytics}. In \bibinfo{booktitle}{\emph{Proceedings of the Third International Conference on Learning Analytics and Knowledge}} (Leuven, Belgium) \emph{(\bibinfo{series}{LAK '13})}. \bibinfo{publisher}{Association for Computing Machinery}, \bibinfo{address}{New York, NY, USA}, \bibinfo{pages}{102--106}.
\newblock
\showISBNx{9781450317856}
\urldef\tempurl%
\url{https://doi.org/10.1145/2460296.2460316}
\showDOI{\tempurl}


\bibitem[Burger et~al\mbox{.}(2018)]%
        {Burger2018-pm}
\bibfield{author}{\bibinfo{person}{Birgitta Burger}, \bibinfo{person}{Anna Puupponen}, {and} \bibinfo{person}{Tommi Jantunen}.} \bibinfo{year}{2018}\natexlab{}.
\newblock \showarticletitle{Synchronizing eye tracking and optical motion capture: How to bring them together}.
\newblock \bibinfo{journal}{\emph{Journal of eye movement research}} \bibinfo{volume}{11}, \bibinfo{number}{2} (\bibinfo{date}{May} \bibinfo{year}{2018}), \bibinfo{pages}{1--16}.
\newblock
\showISSN{1995-8692}
\urldef\tempurl%
\url{https://doi.org/10.16910/jemr.11.2.5}
\showDOI{\tempurl}


\bibitem[Burian(1997)]%
        {Burian1997-pj}
\bibfield{author}{\bibinfo{person}{R~M Burian}.} \bibinfo{year}{1997}\natexlab{}.
\newblock \showarticletitle{Exploratory experimentation and the role of histochemical techniques in the work of Jean Brachet, 1938-1952}.
\newblock \bibinfo{journal}{\emph{History and philosophy of the life sciences}} \bibinfo{volume}{19}, \bibinfo{number}{1} (\bibinfo{year}{1997}), \bibinfo{pages}{27--45}.
\newblock
\showISSN{0391-9714}


\bibitem[Cha and Yu(2022)]%
        {Cha2022-mq}
\bibfield{author}{\bibinfo{person}{Jackie~S Cha} {and} \bibinfo{person}{Denny Yu}.} \bibinfo{year}{2022}\natexlab{}.
\newblock \showarticletitle{Objective Measures of Surgeon {Non-Technical} Skills in Surgery: A Scoping Review}.
\newblock \bibinfo{journal}{\emph{Human factors}} \bibinfo{volume}{64}, \bibinfo{number}{1} (\bibinfo{date}{Feb.} \bibinfo{year}{2022}), \bibinfo{pages}{42--73}.
\newblock
\showISSN{0018-7208, 1547-8181}
\urldef\tempurl%
\url{https://doi.org/10.1177/0018720821995319}
\showDOI{\tempurl}


\bibitem[Chen et~al\mbox{.}(2019)]%
        {Chen2019-nb}
\bibfield{author}{\bibinfo{person}{Richard Chen}, \bibinfo{person}{Filip Jankovic}, \bibinfo{person}{Nikki Marinsek}, \bibinfo{person}{Luca Foschini}, \bibinfo{person}{Lampros Kourtis}, \bibinfo{person}{Alessio Signorini}, \bibinfo{person}{Melissa Pugh}, \bibinfo{person}{Jie Shen}, \bibinfo{person}{Roy Yaari}, \bibinfo{person}{Vera Maljkovic}, \bibinfo{person}{Marc Sunga}, \bibinfo{person}{Han~Hee Song}, \bibinfo{person}{Hyun~Joon Jung}, \bibinfo{person}{Belle Tseng}, {and} \bibinfo{person}{Andrew Trister}.} \bibinfo{year}{2019}\natexlab{}.
\newblock \showarticletitle{Developing Measures of Cognitive Impairment in the Real World from {Consumer-Grade} Multimodal Sensor Streams}. In \bibinfo{booktitle}{\emph{Proceedings of the 25th {ACM} {SIGKDD} International Conference on Knowledge Discovery \& Data Mining}} (Anchorage, AK, USA) \emph{(\bibinfo{series}{KDD '19})}. \bibinfo{publisher}{Association for Computing Machinery}, \bibinfo{address}{New York, NY, USA}, \bibinfo{pages}{2145--2155}.
\newblock
\showISBNx{9781450362016}
\urldef\tempurl%
\url{https://doi.org/10.1145/3292500.3330690}
\showDOI{\tempurl}


\bibitem[Cloude et~al\mbox{.}(2022)]%
        {Cloude2022-zk}
\bibfield{author}{\bibinfo{person}{Elizabeth~B Cloude}, \bibinfo{person}{Megan~D Wiedbusch}, \bibinfo{person}{Daryn~A Dever}, \bibinfo{person}{Dario Torre}, {and} \bibinfo{person}{Roger Azevedo}.} \bibinfo{year}{2022}\natexlab{}.
\newblock \showarticletitle{The Role of Metacognition and Self-regulation on Clinical Reasoning: Leveraging Multimodal Learning Analytics to Transform Medical Education}.
\newblock In \bibinfo{booktitle}{\emph{The Multimodal Learning Analytics Handbook}}, \bibfield{editor}{\bibinfo{person}{Michail Giannakos}, \bibinfo{person}{Daniel Spikol}, \bibinfo{person}{Daniele Di~Mitri}, \bibinfo{person}{Kshitij Sharma}, \bibinfo{person}{Xavier Ochoa}, {and} \bibinfo{person}{Rawad Hammad}} (Eds.). \bibinfo{publisher}{Springer International Publishing}, \bibinfo{address}{Cham}, \bibinfo{pages}{105--129}.
\newblock
\showISBNx{9783031080760}
\urldef\tempurl%
\url{https://doi.org/10.1007/978-3-031-08076-0\_5}
\showDOI{\tempurl}


\bibitem[Conole et~al\mbox{.}(2011)]%
        {Conole2011-zv}
\bibfield{author}{\bibinfo{person}{G Conole}, \bibinfo{person}{D Ga{\v s}evi{\'c}}, \bibinfo{person}{P Long}, {and} \bibinfo{person}{G Siemens}.} \bibinfo{year}{2011}\natexlab{}.
\newblock \showarticletitle{Message from the {LAK} 2011 General \& Program Chairs}. In \bibinfo{booktitle}{\emph{Proceedings of the 1st International Conference on Learning Analytics and Knowledge}}. \bibinfo{publisher}{ACM}, \bibinfo{address}{New York, NY}, \bibinfo{pages}{1--2}.
\newblock


\bibitem[Cui(2019)]%
        {Cui2019-os}
\bibfield{author}{\bibinfo{person}{Wenqiang Cui}.} \bibinfo{year}{2019}\natexlab{}.
\newblock \showarticletitle{Visual Analytics: A Comprehensive Overview}.
\newblock \bibinfo{journal}{\emph{IEEE Access}}  \bibinfo{volume}{7} (\bibinfo{year}{2019}), \bibinfo{pages}{81555--81573}.
\newblock
\showISSN{2169-3536}
\urldef\tempurl%
\url{https://doi.org/10.1109/ACCESS.2019.2923736}
\showDOI{\tempurl}


\bibitem[David et~al\mbox{.}(2022)]%
        {David2022-eg}
\bibfield{author}{\bibinfo{person}{Lida~Z David}, \bibinfo{person}{Maaike~D Endedijk}, {and} \bibinfo{person}{Piet Van~den Bossche}.} \bibinfo{year}{2022}\natexlab{}.
\newblock \showarticletitle{Investigating Interaction Dynamics: A Temporal Approach to Team Learning}.
\newblock In \bibinfo{booktitle}{\emph{Methods for Researching Professional Learning and Development: Challenges, Applications and Empirical Illustrations}}, \bibfield{editor}{\bibinfo{person}{Michael Goller}, \bibinfo{person}{Eva Kyndt}, \bibinfo{person}{Susanna Paloniemi}, {and} \bibinfo{person}{Crina Dam{\c s}a}} (Eds.). \bibinfo{publisher}{Springer International Publishing}, \bibinfo{address}{Cham}, \bibinfo{pages}{187--209}.
\newblock
\showISBNx{9783031085185}
\urldef\tempurl%
\url{https://doi.org/10.1007/978-3-031-08518-5\_9}
\showDOI{\tempurl}


\bibitem[Devezer and Buzbas(2021)]%
        {Devezer2021}
\bibfield{author}{\bibinfo{person}{Berna Devezer} {and} \bibinfo{person}{Erkan Buzbas}.} \bibinfo{year}{2021}\natexlab{}.
\newblock \bibinfo{title}{Minimum Viable Experiment to Replicate}.
\newblock
\newblock
\urldef\tempurl%
\url{http://philsci-archive.pitt.edu/21475/}
\showURL{%
\tempurl}


\bibitem[Di~Mitri et~al\mbox{.}(2018)]%
        {di2018signals}
\bibfield{author}{\bibinfo{person}{Daniele Di~Mitri}, \bibinfo{person}{Jan Schneider}, \bibinfo{person}{Marcus Specht}, {and} \bibinfo{person}{Hendrik Drachsler}.} \bibinfo{year}{2018}\natexlab{}.
\newblock \showarticletitle{From signals to knowledge: A conceptual model for multimodal learning analytics}.
\newblock \bibinfo{journal}{\emph{Journal of Computer Assisted Learning}} \bibinfo{volume}{34}, \bibinfo{number}{4} (\bibinfo{year}{2018}), \bibinfo{pages}{338--349}.
\newblock


\bibitem[Faiz et~al\mbox{.}(2021)]%
        {Faiz2021-le}
\bibfield{author}{\bibinfo{person}{Farina Faiz}, \bibinfo{person}{Yoshinori Ideno}, \bibinfo{person}{Hiromichi Iwasaki}, \bibinfo{person}{Yoko Muroi}, {and} \bibinfo{person}{Sozo Inoue}.} \bibinfo{year}{2021}\natexlab{}.
\newblock \showarticletitle{Multilabel Classification of Nursing Activities in a Realistic Scenario}.
\newblock In \bibinfo{booktitle}{\emph{Activity and Behavior Computing}}, \bibfield{editor}{\bibinfo{person}{Md~Atiqur~Rahman Ahad}, \bibinfo{person}{Sozo Inoue}, \bibinfo{person}{Daniel Roggen}, {and} \bibinfo{person}{Kaori Fujinami}} (Eds.). \bibinfo{publisher}{Springer Singapore}, \bibinfo{address}{Singapore}, \bibinfo{pages}{269--288}.
\newblock
\showISBNx{9789811589447}
\urldef\tempurl%
\url{https://doi.org/10.1007/978-981-15-8944-7\_17}
\showDOI{\tempurl}


\bibitem[Fern{\'a}ndez-M{\'e}ndez et~al\mbox{.}(2019)]%
        {Fernandez-Mendez2019-on}
\bibfield{author}{\bibinfo{person}{Felipe Fern{\'a}ndez-M{\'e}ndez}, \bibinfo{person}{Mart{\'\i}n Otero-Agra}, \bibinfo{person}{Cristian Abelairas-G{\'o}mez}, \bibinfo{person}{Nieves~Mar{\'\i}a S{\'a}ez-Gallego}, \bibinfo{person}{Antonio Rodr{\'\i}guez-N{\'u}{\~n}ez}, {and} \bibinfo{person}{Roberto Barcala-Furelos}.} \bibinfo{year}{2019}\natexlab{}.
\newblock \showarticletitle{{ABCDE} approach to victims by lifeguards: How do they manage a critical patient? A cross sectional simulation study}.
\newblock \bibinfo{journal}{\emph{PloS one}} \bibinfo{volume}{14}, \bibinfo{number}{4} (\bibinfo{date}{April} \bibinfo{year}{2019}), \bibinfo{pages}{e0212080}.
\newblock
\showISSN{1932-6203}
\urldef\tempurl%
\url{https://doi.org/10.1371/journal.pone.0212080}
\showDOI{\tempurl}


\bibitem[Fernandez-Nieto et~al\mbox{.}(2021)]%
        {Fernandez-Nieto2021-fp}
\bibfield{author}{\bibinfo{person}{Gloria Fernandez-Nieto}, \bibinfo{person}{Roberto Martinez-Maldonado}, \bibinfo{person}{Vanessa Echeverria}, \bibinfo{person}{Kirsty Kitto}, \bibinfo{person}{Pengcheng An}, {and} \bibinfo{person}{Simon Buckingham~Shum}.} \bibinfo{year}{2021}\natexlab{}.
\newblock \showarticletitle{What Can Analytics for Teamwork Proxemics Reveal About Positioning Dynamics In Clinical Simulations?}
\newblock \bibinfo{journal}{\emph{Proc. ACM Hum.-Comput. Interact.}} \bibinfo{volume}{5}, \bibinfo{number}{CSCW1} (\bibinfo{date}{April} \bibinfo{year}{2021}), \bibinfo{pages}{1--24}.
\newblock
\urldef\tempurl%
\url{https://doi.org/10.1145/3449284}
\showDOI{\tempurl}


\bibitem[Frank et~al\mbox{.}(2009)]%
        {Frank2009-sn}
\bibfield{author}{\bibinfo{person}{Michael~C Frank}, \bibinfo{person}{Edward Vul}, {and} \bibinfo{person}{Scott~P Johnson}.} \bibinfo{year}{2009}\natexlab{}.
\newblock \showarticletitle{Development of infants' attention to faces during the first year}.
\newblock \bibinfo{journal}{\emph{Cognition}} \bibinfo{volume}{110}, \bibinfo{number}{2} (\bibinfo{date}{Feb.} \bibinfo{year}{2009}), \bibinfo{pages}{160--170}.
\newblock
\showISSN{0010-0277, 1873-7838}
\urldef\tempurl%
\url{https://doi.org/10.1016/j.cognition.2008.11.010}
\showDOI{\tempurl}


\bibitem[Gigerenzer(1991)]%
        {Gigerenzer1991-iu}
\bibfield{author}{\bibinfo{person}{Gerd Gigerenzer}.} \bibinfo{year}{1991}\natexlab{}.
\newblock \showarticletitle{From tools to theories: A heuristic of discovery in cognitive psychology}.
\newblock \bibinfo{journal}{\emph{Psychological review}} \bibinfo{volume}{98}, \bibinfo{number}{2} (\bibinfo{date}{April} \bibinfo{year}{1991}), \bibinfo{pages}{254--267}.
\newblock
\showISSN{0033-295X}
\urldef\tempurl%
\url{https://doi.org/10.1037/0033-295X.98.2.254}
\showDOI{\tempurl}


\bibitem[Gu et~al\mbox{.}(2021)]%
        {Gu2021-nh}
\bibfield{author}{\bibinfo{person}{Zhenyu Gu}, \bibinfo{person}{Chenhao Jin}, \bibinfo{person}{Danny Chang}, {and} \bibinfo{person}{Liqun Zhang}.} \bibinfo{year}{2021}\natexlab{}.
\newblock \showarticletitle{Predicting webpage aesthetics with heatmap entropy}.
\newblock \bibinfo{journal}{\emph{Behaviour \& information technology}} \bibinfo{volume}{40}, \bibinfo{number}{7} (\bibinfo{date}{May} \bibinfo{year}{2021}), \bibinfo{pages}{676--690}.
\newblock
\showISSN{0144-929X}
\urldef\tempurl%
\url{https://doi.org/10.1080/0144929X.2020.1717626}
\showDOI{\tempurl}


\bibitem[Heilala et~al\mbox{.}(2022)]%
        {Heilala2022-xt}
\bibfield{author}{\bibinfo{person}{Ville Heilala}, \bibinfo{person}{P{\"a}ivikki J{\"a}{\"a}skel{\"a}}, \bibinfo{person}{Mirka Saarela}, \bibinfo{person}{Anna-Stina Kuula}, \bibinfo{person}{Anne Eskola}, {and} \bibinfo{person}{Tommi K{\"a}rkk{\"a}inen}.} \bibinfo{year}{2022}\natexlab{}.
\newblock \showarticletitle{``Sitting at the Stern and Holding the Rudder'': Teachers' Reflections on Action in Higher Education Based on Student Agency Analytics}.
\newblock In \bibinfo{booktitle}{\emph{Digital Teaching and Learning in Higher Education: Developing and Disseminating Skills for Blended Learning}}, \bibfield{editor}{\bibinfo{person}{Leonid Chechurin}} (Ed.). \bibinfo{publisher}{Palgrave Macmillan}, \bibinfo{address}{Cham}, \bibinfo{pages}{71--91}.
\newblock
\showISBNx{9783031008016}
\urldef\tempurl%
\url{https://doi.org/10.1007/978-3-031-00801-6\_4}
\showDOI{\tempurl}


\bibitem[Holmqvist et~al\mbox{.}(2011)]%
        {Holmqvist2011-tr}
\bibfield{author}{\bibinfo{person}{Kenneth Holmqvist}, \bibinfo{person}{Marcus Nystr{\"o}m}, \bibinfo{person}{Richard Andersson}, \bibinfo{person}{Richard Dewhurst}, \bibinfo{person}{Halszka Jarodzka}, {and} \bibinfo{person}{Joost van~de Weijer}.} \bibinfo{year}{2011}\natexlab{}.
\newblock \bibinfo{booktitle}{\emph{Eye Tracking: A comprehensive guide to methods and measures}}.
\newblock \bibinfo{publisher}{OUP Oxford}, \bibinfo{address}{Oxford}.
\newblock
\showISBNx{9780191625428}


\bibitem[Hutchins(1995)]%
        {Hutchins1995-bp}
\bibfield{author}{\bibinfo{person}{Edwin Hutchins}.} \bibinfo{year}{1995}\natexlab{}.
\newblock \bibinfo{booktitle}{\emph{Cognition in the Wild}}.
\newblock \bibinfo{publisher}{MIT Press}, \bibinfo{address}{Cambridge, MA}.
\newblock
\showISBNx{9780262082310}


\bibitem[Karaim et~al\mbox{.}(2019)]%
        {Karaim2019-yn}
\bibfield{author}{\bibinfo{person}{Malek Karaim}, \bibinfo{person}{Aboelmagd Noureldin}, {and} \bibinfo{person}{Tashfeen~B Karamat}.} \bibinfo{year}{2019}\natexlab{}.
\newblock \showarticletitle{Low-cost {IMU} Data Denoising using {Savitzky-Golay} Filters}. In \bibinfo{booktitle}{\emph{2019 International Conference on Communications, Signal Processing, and their Applications ({ICCSPA})}}. \bibinfo{publisher}{IEEE}, \bibinfo{address}{Piscataway}, \bibinfo{pages}{1--5}.
\newblock
\urldef\tempurl%
\url{https://doi.org/10.1109/ICCSPA.2019.8713728}
\showDOI{\tempurl}


\bibitem[Keim(2001)]%
        {Keim2001-og}
\bibfield{author}{\bibinfo{person}{Daniel~A Keim}.} \bibinfo{year}{2001}\natexlab{}.
\newblock \showarticletitle{Visual exploration of large data sets}.
\newblock \bibinfo{journal}{\emph{Commun. ACM}} \bibinfo{volume}{44}, \bibinfo{number}{8} (\bibinfo{date}{Aug.} \bibinfo{year}{2001}), \bibinfo{pages}{38--44}.
\newblock
\showISSN{0001-0782}
\urldef\tempurl%
\url{https://doi.org/10.1145/381641.381656}
\showDOI{\tempurl}


\bibitem[Keim(2002)]%
        {Keim2002-jt}
\bibfield{author}{\bibinfo{person}{Daniel~A Keim}.} \bibinfo{year}{2002}\natexlab{}.
\newblock \showarticletitle{Information visualization and visual data mining}.
\newblock \bibinfo{journal}{\emph{IEEE transactions on visualization and computer graphics}} \bibinfo{volume}{8}, \bibinfo{number}{1} (\bibinfo{date}{Jan.} \bibinfo{year}{2002}), \bibinfo{pages}{1--8}.
\newblock
\showISSN{1077-2626, 1941-0506}
\urldef\tempurl%
\url{https://doi.org/10.1109/2945.981847}
\showDOI{\tempurl}


\bibitem[Kim et~al\mbox{.}(2015)]%
        {Kim2015-qa}
\bibfield{author}{\bibinfo{person}{Dae-Yeob Kim}, \bibinfo{person}{Soo-Hyung Kim}, \bibinfo{person}{Daeseon Choi}, {and} \bibinfo{person}{Seung-Hun Jin}.} \bibinfo{year}{2015}\natexlab{}.
\newblock \showarticletitle{Accurate Indoor Proximity Zone Detection Based on Time Window and Frequency with Bluetooth Low Energy}.
\newblock \bibinfo{journal}{\emph{Procedia computer science}}  \bibinfo{volume}{56} (\bibinfo{date}{Jan.} \bibinfo{year}{2015}), \bibinfo{pages}{88--95}.
\newblock
\showISSN{1877-0509}
\urldef\tempurl%
\url{https://doi.org/10.1016/j.procs.2015.07.199}
\showDOI{\tempurl}


\bibitem[Kolbe and Boos(2019)]%
        {Kolbe2019-ym}
\bibfield{author}{\bibinfo{person}{Michaela Kolbe} {and} \bibinfo{person}{Margarete Boos}.} \bibinfo{year}{2019}\natexlab{}.
\newblock \showarticletitle{Laborious but Elaborate: The Benefits of Really Studying Team Dynamics}.
\newblock \bibinfo{journal}{\emph{Frontiers in psychology}}  \bibinfo{volume}{10} (\bibinfo{date}{June} \bibinfo{year}{2019}), \bibinfo{pages}{1478}.
\newblock
\showISSN{1664-1078}
\urldef\tempurl%
\url{https://doi.org/10.3389/fpsyg.2019.01478}
\showDOI{\tempurl}


\bibitem[L{\"a}ms{\"a} et~al\mbox{.}(2023)]%
        {Lamsa2023-al}
\bibfield{author}{\bibinfo{person}{Joni L{\"a}ms{\"a}}, \bibinfo{person}{Joonas Mannonen}, \bibinfo{person}{Ari Tuhkala}, \bibinfo{person}{Ville Heilala}, \bibinfo{person}{Arto Helovuo}, \bibinfo{person}{Ilkka Tynkkynen}, \bibinfo{person}{Emilia Lampi}, \bibinfo{person}{Katriina Sipil{\"a}inen}, \bibinfo{person}{Tommi K{\"a}rkk{\"a}inen}, {and} \bibinfo{person}{Raija H{\"a}m{\"a}l{\"a}inen}.} \bibinfo{year}{2023}\natexlab{}.
\newblock \showarticletitle{Capturing cognitive load management during authentic virtual reality flight training with behavioural and physiological indicators}.
\newblock \bibinfo{journal}{\emph{Journal of computer assisted learning}} \bibinfo{volume}{39}, \bibinfo{number}{5} (\bibinfo{date}{April} \bibinfo{year}{2023}), \bibinfo{pages}{1553--1563}.
\newblock
\showISSN{0266-4909, 1365-2729}
\urldef\tempurl%
\url{https://doi.org/10.1111/jcal.12817}
\showDOI{\tempurl}


\bibitem[Lanini-Maggi et~al\mbox{.}(2021)]%
        {Lanini-Maggi2021-vj}
\bibfield{author}{\bibinfo{person}{Sara Lanini-Maggi}, \bibinfo{person}{Ian~T Ruginski}, \bibinfo{person}{Thomas~F Shipley}, \bibinfo{person}{Christophe Hurter}, \bibinfo{person}{Andrew~T Duchowski}, \bibinfo{person}{Benny~B Briesemeister}, \bibinfo{person}{Jihyun Lee}, {and} \bibinfo{person}{Sara~I Fabrikant}.} \bibinfo{year}{2021}\natexlab{}.
\newblock \showarticletitle{Assessing how visual search entropy and engagement predict performance in a multiple-objects tracking air traffic control task}.
\newblock \bibinfo{journal}{\emph{Computers in Human Behavior Reports}}  \bibinfo{volume}{4} (\bibinfo{date}{Aug.} \bibinfo{year}{2021}), \bibinfo{pages}{100127}.
\newblock
\showISSN{2451-9588}
\urldef\tempurl%
\url{https://doi.org/10.1016/j.chbr.2021.100127}
\showDOI{\tempurl}


\bibitem[Lee et~al\mbox{.}(2019)]%
        {Lee2019-cq}
\bibfield{author}{\bibinfo{person}{Joy~Yeonjoo Lee}, \bibinfo{person}{Jeroen Donkers}, \bibinfo{person}{Halszka Jarodzka}, {and} \bibinfo{person}{Jeroen J~G van Merri{\"e}nboer}.} \bibinfo{year}{2019}\natexlab{}.
\newblock \showarticletitle{How prior knowledge affects problem-solving performance in a medical simulation game: Using game-logs and eye-tracking}.
\newblock \bibinfo{journal}{\emph{Computers in human behavior}}  \bibinfo{volume}{99} (\bibinfo{date}{Oct.} \bibinfo{year}{2019}), \bibinfo{pages}{268--277}.
\newblock
\showISSN{0747-5632}
\urldef\tempurl%
\url{https://doi.org/10.1016/j.chb.2019.05.035}
\showDOI{\tempurl}


\bibitem[Li et~al\mbox{.}(2021)]%
        {Li2021-fk}
\bibfield{author}{\bibinfo{person}{Maximilian~Xiling Li}, \bibinfo{person}{Mario Nadj}, \bibinfo{person}{Alexander Maedche}, \bibinfo{person}{Dirk Ifenthaler}, {and} \bibinfo{person}{Johannes W{\"o}hler}.} \bibinfo{year}{2021}\natexlab{}.
\newblock \showarticletitle{Towards a physiological computing infrastructure for researching students' flow in remote learning}.
\newblock \bibinfo{journal}{\emph{Technology Knowledge and Learning}} \bibinfo{volume}{27}, \bibinfo{number}{1} (\bibinfo{date}{Sept.} \bibinfo{year}{2021}), \bibinfo{pages}{365--384}.
\newblock
\showISSN{2211-1662, 2211-1670}
\urldef\tempurl%
\url{https://doi.org/10.1007/s10758-021-09569-4}
\showDOI{\tempurl}


\bibitem[Linja et~al\mbox{.}(2023)]%
        {Linja2023feature}
\bibfield{author}{\bibinfo{person}{Joakim Linja}, \bibinfo{person}{Joonas H{\" a}m{\" a}l{\" a}inen}, \bibinfo{person}{Paavo Nieminen}, {and} \bibinfo{person}{Tommi K{\" a}rkk{\" a}inen}.} \bibinfo{year}{2023}\natexlab{}.
\newblock \showarticletitle{Feature selection for distance-based regression: An umbrella review and a one-shot wrapper}.
\newblock \bibinfo{journal}{\emph{Neurocomputing}}  \bibinfo{volume}{518} (\bibinfo{year}{2023}), \bibinfo{pages}{344--359}.
\newblock


\bibitem[MacKay(2003)]%
        {MacKay2003-qt}
\bibfield{author}{\bibinfo{person}{David J~C MacKay}.} \bibinfo{year}{2003}\natexlab{}.
\newblock \bibinfo{booktitle}{\emph{Information Theory, Inference and Learning Algorithms}}.
\newblock \bibinfo{publisher}{Cambridge University Press}, \bibinfo{address}{Cambridge}.
\newblock
\showISBNx{9780521642989}


\bibitem[Martinez-Maldonado et~al\mbox{.}(2023)]%
        {martinez2023lessons}
\bibfield{author}{\bibinfo{person}{Roberto Martinez-Maldonado}, \bibinfo{person}{Vanessa Echeverria}, \bibinfo{person}{Gloria Fernandez-Nieto}, \bibinfo{person}{Lixiang Yan}, \bibinfo{person}{Linxuan Zhao}, \bibinfo{person}{Riordan Alfredo}, \bibinfo{person}{Xinyu Li}, \bibinfo{person}{Samantha Dix}, \bibinfo{person}{Hollie Jaggard}, \bibinfo{person}{Rosie Wotherspoon}, \bibinfo{person}{Abra Osborne}, \bibinfo{person}{Simon~Buckingham Shum}, {and} \bibinfo{person}{Dragan Ga{\v s}evi{\'c}}.} \bibinfo{year}{2023}\natexlab{}.
\newblock \showarticletitle{Lessons Learnt from a Multimodal Learning Analytics Deployment In-the-wild}.
\newblock \bibinfo{journal}{\emph{ACM Trans. Comput.-Hum. Interact.}} \bibinfo{volume}{31}, \bibinfo{number}{1} (\bibinfo{date}{Sept.} \bibinfo{year}{2023}), \bibinfo{pages}{1--41}.
\newblock
\showISSN{1073-0516}
\urldef\tempurl%
\url{https://doi.org/10.1145/3622784}
\showDOI{\tempurl}


\bibitem[Martinez‐Maldonado et~al\mbox{.}(2020)]%
        {MartinezMaldonado2020-pa}
\bibfield{author}{\bibinfo{person}{Roberto Martinez‐Maldonado}, \bibinfo{person}{Jurgen Schulte}, \bibinfo{person}{Vanessa Echeverria}, \bibinfo{person}{Yuveena Gopalan}, {and} \bibinfo{person}{Simon~Buckingham Shum}.} \bibinfo{year}{2020}\natexlab{}.
\newblock \showarticletitle{Where is the teacher? Digital analytics for classroom proxemics}.
\newblock \bibinfo{journal}{\emph{Journal of Computer Assisted Learning}}  \bibinfo{volume}{16} (\bibinfo{date}{May} \bibinfo{year}{2020}), \bibinfo{pages}{1}.
\newblock
\showISSN{0266-4909, 1365-2729}
\urldef\tempurl%
\url{https://doi.org/10.1111/jcal.12444}
\showDOI{\tempurl}


\bibitem[Molenaar et~al\mbox{.}(2023)]%
        {Molenaar2023-ug}
\bibfield{author}{\bibinfo{person}{Inge Molenaar}, \bibinfo{person}{Susanne~de Mooij}, \bibinfo{person}{Roger Azevedo}, \bibinfo{person}{Maria Bannert}, \bibinfo{person}{Sanna J{\"a}rvel{\"a}}, {and} \bibinfo{person}{Dragan Ga{\v s}evi{\'c}}.} \bibinfo{year}{2023}\natexlab{}.
\newblock \showarticletitle{Measuring self-regulated learning and the role of {AI}: Five years of research using multimodal multichannel data}.
\newblock \bibinfo{journal}{\emph{Computers in human behavior}}  \bibinfo{volume}{139} (\bibinfo{date}{Feb.} \bibinfo{year}{2023}), \bibinfo{pages}{107540}.
\newblock
\showISSN{0747-5632}
\urldef\tempurl%
\url{https://doi.org/10.1016/j.chb.2022.107540}
\showDOI{\tempurl}


\bibitem[Momen and Fernie(2010)]%
        {Momen2010-jt}
\bibfield{author}{\bibinfo{person}{Kaveh Momen} {and} \bibinfo{person}{Geoff~R Fernie}.} \bibinfo{year}{2010}\natexlab{}.
\newblock \showarticletitle{Nursing activity recognition using an inexpensive game controller: An application to infection control}.
\newblock \bibinfo{journal}{\emph{Technology and health care: official journal of the European Society for Engineering and Medicine}} \bibinfo{volume}{18}, \bibinfo{number}{6} (\bibinfo{year}{2010}), \bibinfo{pages}{393--408}.
\newblock
\showISSN{0928-7329, 1878-7401}
\urldef\tempurl%
\url{https://doi.org/10.3233/THC-2010-0600}
\showDOI{\tempurl}


\bibitem[Morita et~al\mbox{.}(2022)]%
        {Morita2022-kq}
\bibfield{author}{\bibinfo{person}{Akio Morita}, \bibinfo{person}{Yasuo Murai}, {and} \bibinfo{person}{Mamoru Mitsuishi}.} \bibinfo{year}{2022}\natexlab{}.
\newblock \showarticletitle{Current and Future Microsurgical Skills Assessment}.
\newblock In \bibinfo{booktitle}{\emph{Learning and Career Development in Neurosurgery: {Values-Based} Medical Education}}, \bibfield{editor}{\bibinfo{person}{Ahmed Ammar}} (Ed.). \bibinfo{publisher}{Springer International Publishing}, \bibinfo{address}{Cham}, \bibinfo{pages}{349--356}.
\newblock
\showISBNx{9783031020780}
\urldef\tempurl%
\url{https://doi.org/10.1007/978-3-031-02078-0\_30}
\showDOI{\tempurl}


\bibitem[Ouhaichi et~al\mbox{.}(2023)]%
        {ouhaichi2023research}
\bibfield{author}{\bibinfo{person}{Hamza Ouhaichi}, \bibinfo{person}{Daniel Spikol}, {and} \bibinfo{person}{Bahtijar Vogel}.} \bibinfo{year}{2023}\natexlab{}.
\newblock \showarticletitle{Research trends in multimodal learning analytics: A systematic mapping study}.
\newblock \bibinfo{journal}{\emph{Computers and Education: Artificial Intelligence}}  \bibinfo{volume}{4} (\bibinfo{date}{Jan.} \bibinfo{year}{2023}), \bibinfo{pages}{100136}.
\newblock
\showISSN{2666-920X}
\urldef\tempurl%
\url{https://doi.org/10.1016/j.caeai.2023.100136}
\showDOI{\tempurl}


\bibitem[Schafer(2011)]%
        {Schafer2011-nx}
\bibfield{author}{\bibinfo{person}{Ronald~W Schafer}.} \bibinfo{year}{2011}\natexlab{}.
\newblock \showarticletitle{What Is a {Savitzky-Golay} Filter? [Lecture Notes]}.
\newblock \bibinfo{journal}{\emph{IEEE Signal Processing Magazine}} \bibinfo{volume}{28}, \bibinfo{number}{4} (\bibinfo{date}{July} \bibinfo{year}{2011}), \bibinfo{pages}{111--117}.
\newblock
\showISSN{1558-0792}
\urldef\tempurl%
\url{https://doi.org/10.1109/MSP.2011.941097}
\showDOI{\tempurl}


\bibitem[Schoeber et~al\mbox{.}(2022)]%
        {Schoeber2022-og}
\bibfield{author}{\bibinfo{person}{Nino H~C Schoeber}, \bibinfo{person}{Marjolein Linders}, \bibinfo{person}{Mathijs Binkhorst}, \bibinfo{person}{Willem-Pieter De~Boode}, \bibinfo{person}{Jos M~T Draaisma}, \bibinfo{person}{Marlies Morsink}, \bibinfo{person}{Anneliese Nusmeier}, \bibinfo{person}{Martijn Pas}, \bibinfo{person}{Christine van Riessen}, \bibinfo{person}{Nigel~M Turner}, \bibinfo{person}{Rutger Verhage}, \bibinfo{person}{Cornelia R M~G Fluit}, {and} \bibinfo{person}{Marije Hogeveen}.} \bibinfo{year}{2022}\natexlab{}.
\newblock \showarticletitle{Healthcare professionals' knowledge of the systematic {ABCDE} approach: a cross-sectional study}.
\newblock \bibinfo{journal}{\emph{BMC emergency medicine}} \bibinfo{volume}{22}, \bibinfo{number}{1} (\bibinfo{date}{Dec.} \bibinfo{year}{2022}), \bibinfo{pages}{202}.
\newblock
\showISSN{1471-227X}
\urldef\tempurl%
\url{https://doi.org/10.1186/s12873-022-00753-y}
\showDOI{\tempurl}


\bibitem[Shannon(1948)]%
        {Shannon1948-gy}
\bibfield{author}{\bibinfo{person}{C~E Shannon}.} \bibinfo{year}{1948}\natexlab{}.
\newblock \showarticletitle{A mathematical theory of communication}.
\newblock \bibinfo{journal}{\emph{The Bell system technical journal}} \bibinfo{volume}{27}, \bibinfo{number}{3} (\bibinfo{year}{1948}), \bibinfo{pages}{379--423}.
\newblock
\showISSN{1538-7305}
\urldef\tempurl%
\url{https://doi.org/10.1145/584091.584093}
\showDOI{\tempurl}


\bibitem[Shiferaw et~al\mbox{.}(2019)]%
        {Shiferaw2019-ld}
\bibfield{author}{\bibinfo{person}{Brook Shiferaw}, \bibinfo{person}{Luke Downey}, {and} \bibinfo{person}{David Crewther}.} \bibinfo{year}{2019}\natexlab{}.
\newblock \showarticletitle{A review of gaze entropy as a measure of visual scanning efficiency}.
\newblock \bibinfo{journal}{\emph{Neuroscience and biobehavioral reviews}}  \bibinfo{volume}{96} (\bibinfo{date}{Jan.} \bibinfo{year}{2019}), \bibinfo{pages}{353--366}.
\newblock
\showISSN{0149-7634, 1873-7528}
\urldef\tempurl%
\url{https://doi.org/10.1016/j.neubiorev.2018.12.007}
\showDOI{\tempurl}


\bibitem[Steinle(1997)]%
        {Steinle1997-jw}
\bibfield{author}{\bibinfo{person}{Friedrich Steinle}.} \bibinfo{year}{1997}\natexlab{}.
\newblock \showarticletitle{Entering New Fields: Exploratory Uses of Experimentation}.
\newblock \bibinfo{journal}{\emph{Philosophy of science}} \bibinfo{volume}{64}, \bibinfo{number}{S4} (\bibinfo{year}{1997}), \bibinfo{pages}{S65--S74}.
\newblock
\showISSN{0031-8248, 1539-767X}
\urldef\tempurl%
\url{https://doi.org/10.1086/392587}
\showDOI{\tempurl}


\bibitem[Thim et~al\mbox{.}(2012)]%
        {Thim2012-cg}
\bibfield{author}{\bibinfo{person}{Troels Thim}, \bibinfo{person}{Niels Henrik~Vinther Krarup}, \bibinfo{person}{Erik~Lerkevang Grove}, \bibinfo{person}{Claus~Valter Rohde}, {and} \bibinfo{person}{Bo L{\o}fgren}.} \bibinfo{year}{2012}\natexlab{}.
\newblock \showarticletitle{Initial assessment and treatment with the Airway, Breathing, Circulation, Disability, Exposure ({ABCDE}) approach}.
\newblock \bibinfo{journal}{\emph{International journal of general medicine}}  \bibinfo{volume}{5} (\bibinfo{date}{Jan.} \bibinfo{year}{2012}), \bibinfo{pages}{117--121}.
\newblock
\showISSN{1178-7074}
\urldef\tempurl%
\url{https://doi.org/10.2147/IJGM.S28478}
\showDOI{\tempurl}


\bibitem[Tokuno et~al\mbox{.}(2023)]%
        {Tokuno2023-fj}
\bibfield{author}{\bibinfo{person}{Junko Tokuno}, \bibinfo{person}{Tamara~E Carver}, {and} \bibinfo{person}{Gerald~M Fried}.} \bibinfo{year}{2023}\natexlab{}.
\newblock \showarticletitle{Measurement and Management of Cognitive Load in Surgical Education: A Narrative Review}.
\newblock \bibinfo{journal}{\emph{Journal of surgical education}} \bibinfo{volume}{80}, \bibinfo{number}{2} (\bibinfo{date}{Feb.} \bibinfo{year}{2023}), \bibinfo{pages}{208--215}.
\newblock
\showISSN{1931-7204, 1878-7452}
\urldef\tempurl%
\url{https://doi.org/10.1016/j.jsurg.2022.10.001}
\showDOI{\tempurl}


\bibitem[Winne(2019)]%
        {Winne2019-wj}
\bibfield{author}{\bibinfo{person}{Philip~H Winne}.} \bibinfo{year}{2019}\natexlab{}.
\newblock \showarticletitle{Paradigmatic Dimensions of Instrumentation and Analytic Methods in Research on {Self-Regulated} Learning}.
\newblock \bibinfo{journal}{\emph{Computers in human behavior}}  \bibinfo{volume}{96} (\bibinfo{date}{July} \bibinfo{year}{2019}), \bibinfo{pages}{285--289}.
\newblock
\showISSN{0747-5632}
\urldef\tempurl%
\url{https://doi.org/10.1016/j.chb.2019.03.026}
\showDOI{\tempurl}


\bibitem[Worsley et~al\mbox{.}(2021)]%
        {worsley2021new}
\bibfield{author}{\bibinfo{person}{Marcelo Worsley}, \bibinfo{person}{Roberto Martinez-Maldonado}, {and} \bibinfo{person}{Cynthia D'Angelo}.} \bibinfo{year}{2021}\natexlab{}.
\newblock \showarticletitle{A New Era in Multimodal Learning Analytics: Twelve Core Commitments to Ground and Grow MMLA.}
\newblock \bibinfo{journal}{\emph{Journal of Learning Analytics}} \bibinfo{volume}{8}, \bibinfo{number}{3} (\bibinfo{year}{2021}), \bibinfo{pages}{10--27}.
\newblock


\bibitem[Wright et~al\mbox{.}(2022)]%
        {Andrew_Wright2022-yw}
\bibfield{author}{\bibinfo{person}{Andrew~G Wright}, \bibinfo{person}{Rahool Patel}, \bibinfo{person}{Koraly Perez-Edgar}, \bibinfo{person}{Xiaoxue Fu}, \bibinfo{person}{Kayla Brown}, \bibinfo{person}{Sanjib Adhikary}, {and} \bibinfo{person}{Adrian Zurca}.} \bibinfo{year}{2022}\natexlab{}.
\newblock \showarticletitle{{Eye-Tracking} Technology to Determine Procedural Proficiency in {Ultrasound-Guided} Regional Anesthesia}.
\newblock \bibinfo{journal}{\emph{The journal of education in perioperative medicine : JEPM}} \bibinfo{volume}{24}, \bibinfo{number}{1} (\bibinfo{date}{Jan.} \bibinfo{year}{2022}), \bibinfo{pages}{E684}.
\newblock
\showISSN{2333-0406}
\urldef\tempurl%
\url{https://doi.org/10.46374/volxxiv\_issue1\_zurca}
\showDOI{\tempurl}


\bibitem[Yadegaridehkordi et~al\mbox{.}(2019)]%
        {Yadegaridehkordi2019-sz}
\bibfield{author}{\bibinfo{person}{Elaheh Yadegaridehkordi}, \bibinfo{person}{Nurul Fazmidar Binti~Mohd Noor}, \bibinfo{person}{Mohamad Nizam~Bin Ayub}, \bibinfo{person}{Hannyzzura~Binti Affal}, {and} \bibinfo{person}{Nornazlita~Binti Hussin}.} \bibinfo{year}{2019}\natexlab{}.
\newblock \showarticletitle{Affective computing in education: A systematic review and future research}.
\newblock \bibinfo{journal}{\emph{Computers \& education}}  \bibinfo{volume}{142} (\bibinfo{date}{Dec.} \bibinfo{year}{2019}), \bibinfo{pages}{103649}.
\newblock
\showISSN{0360-1315}
\urldef\tempurl%
\url{https://doi.org/10.1016/j.compedu.2019.103649}
\showDOI{\tempurl}


\bibitem[Yan et~al\mbox{.}(2022)]%
        {Yan2022-vo}
\bibfield{author}{\bibinfo{person}{Lixiang Yan}, \bibinfo{person}{Linxuan Zhao}, \bibinfo{person}{Dragan Gasevic}, {and} \bibinfo{person}{Roberto Martinez-Maldonado}.} \bibinfo{year}{2022}\natexlab{}.
\newblock \showarticletitle{Scalability, Sustainability, and Ethicality of Multimodal Learning Analytics}. In \bibinfo{booktitle}{\emph{{LAK22}: 12th International Learning Analytics and Knowledge Conference}} (Online, USA) \emph{(\bibinfo{series}{LAK22})}. \bibinfo{publisher}{Association for Computing Machinery}, \bibinfo{address}{New York, NY, USA}, \bibinfo{pages}{13--23}.
\newblock
\showISBNx{9781450395731}
\urldef\tempurl%
\url{https://doi.org/10.1145/3506860.3506862}
\showDOI{\tempurl}


\bibitem[Zhao et~al\mbox{.}(2023)]%
        {Zhao2023-lx}
\bibfield{author}{\bibinfo{person}{Linxuan Zhao}, \bibinfo{person}{Zachari Swiecki}, \bibinfo{person}{Dragan Gasevic}, \bibinfo{person}{Lixiang Yan}, \bibinfo{person}{Samantha Dix}, \bibinfo{person}{Hollie Jaggard}, \bibinfo{person}{Rosie Wotherspoon}, \bibinfo{person}{Abra Osborne}, \bibinfo{person}{Xinyu Li}, \bibinfo{person}{Riordan Alfredo}, {and} \bibinfo{person}{Roberto Martinez-Maldonado}.} \bibinfo{year}{2023}\natexlab{}.
\newblock \showarticletitle{{METS}: Multimodal Learning Analytics of Embodied Teamwork Learning}. In \bibinfo{booktitle}{\emph{{LAK23}: 13th International Learning Analytics and Knowledge Conference}} (Arlington, TX, USA) \emph{(\bibinfo{series}{LAK2023})}. \bibinfo{publisher}{Association for Computing Machinery}, \bibinfo{address}{New York, NY, USA}, \bibinfo{pages}{186--196}.
\newblock
\showISBNx{9781450398657}
\urldef\tempurl%
\url{https://doi.org/10.1145/3576050.3576076}
\showDOI{\tempurl}


\end{thebibliography}

\end{document}